\documentclass{article}

\usepackage{microtype}
\usepackage{graphicx}
\usepackage{booktabs} 

\usepackage[hidelinks]{hyperref}

\usepackage{fullpage}

\usepackage{amsmath}
\usepackage{amssymb}
\usepackage{mathtools}
\usepackage{amsthm}

\usepackage[capitalize,noabbrev]{cleveref}

\theoremstyle{plain}

\theoremstyle{definition}

\theoremstyle{remark}

\usepackage[textsize=tiny]{todonotes}

\usepackage{graphicx}
\usepackage{bm}
\usepackage{times}
\usepackage{amsmath}
\usepackage{amssymb}
\usepackage{natbib}
\usepackage{multirow}

\usepackage[utf8]{inputenc} 
\usepackage[T1]{fontenc}    
\usepackage{url}            
\usepackage{booktabs}       
\usepackage{amsfonts}       
\usepackage{nicefrac}       
\usepackage{microtype}      
\usepackage{dsfont}
\usepackage{amsthm}
\usepackage{thmtools}
\usepackage{subcaption}
\usepackage{wrapfig}
\usepackage{colortbl}
\usepackage{bm}
\usepackage{pifont}
\usepackage{threeparttable}
\usepackage{pgfplotstable}
\usepackage{thm-restate}

\usepackage{listings}
\definecolor{codegreen}{rgb}{0,0.6,0}
\definecolor{codegray}{rgb}{0.5,0.5,0.5}
\definecolor{codepurple}{rgb}{0.58,0,0.82}
\definecolor{backcolour}{rgb}{0.95,0.95,0.92}

\lstdefinestyle{mystyle}{
    backgroundcolor=\color{backcolour},   
    commentstyle=\color{codegreen},
    keywordstyle=\color{magenta},
    numberstyle=\tiny\color{codegray},
    stringstyle=\color{codepurple},
    basicstyle=\ttfamily\footnotesize,
    breakatwhitespace=false,         
    breaklines=true,                 
    captionpos=b,                    
    keepspaces=true,                 
    numbers=left,                    
    numbersep=5pt,                  
    showspaces=false,                
    showstringspaces=false,
    showtabs=false,                  
    tabsize=2
}

\def \xv {{\bm x}}
\def \yv {{\bm y}}

\def \thetav {{\bm \theta}}
\def \Hv {{\bm H}}

\usepackage{xcolor}

\usepackage{enumitem}

\usepackage[textsize=tiny]{todonotes}

\newcommand{\Ec}{\mathcal{E}}

\newcommand{\Mc}{\mathcal{M}}

\newcommand{\Sc}{\mathcal{S}}
\newcommand{\Tc}{\mathcal{T}}

\newcommand{\Eb}{\mathbb{E}}

\begin{document}

\twocolumn[
\vspace{-2em}
\title{\bf Cross Domain Generative Augmentation: \\ Domain Generalization with Latent Diffusion Models}

\newcommand{\spaces}{{\;\;\;\;\;\;\;\;\;}}
\author{\bf Sobhan Hemati\spaces Mahdi Beitollahi\spaces Amir Hossein Estiri\spaces Bassel Al Omari \\
\bf Xi Chen\spaces Guojun Zhang\\ \\
Huawei Noah's Ark Lab\\ \\
\begin{minipage}{1\linewidth}\centering\small\rm
$\left\{\genfrac{}{}{0pt}{}{\displaystyle\texttt{sobhan.hemati, mahdi.beitollahi, amir.hossein.estiri, bassel.al.omari,}}{\displaystyle\texttt{xi.chen4, guojun.zhang}}\right\}\texttt{@huawei.com}$
\end{minipage}
}
\date{}
\maketitle
]

\begin{abstract}
Despite the huge effort in developing novel regularizers for Domain Generalization (DG), adding simple data augmentation to the vanilla ERM which is a practical implementation of the Vicinal Risk Minimization principle (VRM) \citep{chapelle2000vicinal} outperforms or stays competitive with many of the proposed regularizers. The VRM reduces the estimation error in ERM by replacing the point-wise kernel estimates with a more precise estimation of true data distribution that reduces the gap between data points \textbf{within each domain}. However, in the DG setting, the estimation error of true data distribution by ERM is mainly caused by the distribution shift \textbf{between domains} which cannot be fully addressed by simple data augmentation techniques within each domain. Inspired by this limitation of VRM, we propose a novel data augmentation named Cross Domain Generative Augmentation (CDGA) that replaces the pointwise kernel estimates in ERM with new density estimates in the \textbf{vicinity of domain pairs} so that the gap between domains is further reduced. To this end, CDGA, which is built upon latent diffusion models (LDM), generates synthetic images to fill the gap between all domains and as a result, reduces the non-iidness. We show that CDGA outperforms SOTA DG methods under the Domainbed benchmark. To explain the effectiveness of CDGA, we generate more than 5 Million synthetic images and perform extensive ablation studies including data scaling laws, distribution visualization, domain shift quantification, adversarial robustness, and loss landscape analysis.
\end{abstract}

\section{Introduction}
\label{sec:intro}
Out-of-distribution (OOD) generalization is a critical ability of deep learning models in real-world scenarios, as the target set does not necessarily follow the same distribution as the training set we collect. Unfortunately, under domain shift, i.e., non-iidness between training domains and target domain, the models trained with Empirical Risk Minimization (ERM) \cite{vapnik1999overview} fail to retain their performance in the target domain.

\begin{figure}[ht]
\centering
\includegraphics[width=0.99\columnwidth]{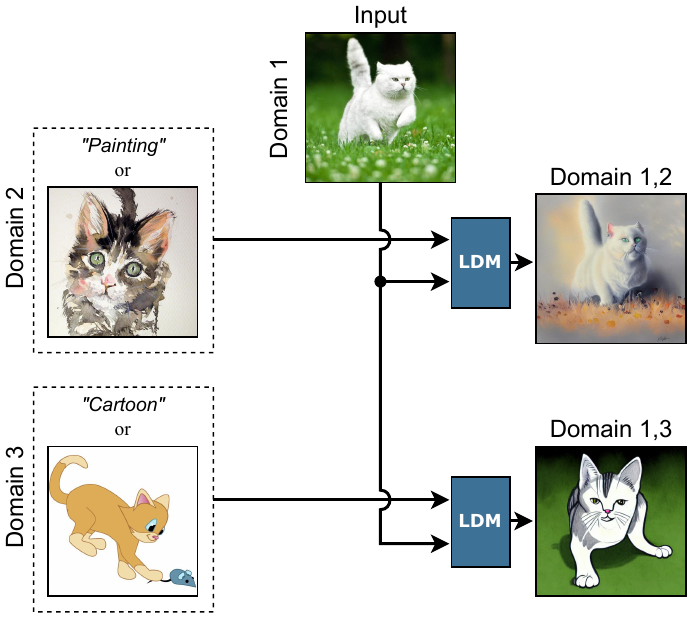} 
\caption{Illustration of cross domain generative augmentation. For each input image of a domain, we generate a new image using the image or the description of another domain.}
\label{fig:illustration}
\end{figure}

The common setting to study OOD generalization is called \emph{domain generalization} (DG) \citet{blanchard2011generalizing}), where multiple source domains are available and the goal is to generalize to an unseen target domain. To improve the OOD generalization of ERM in DG setup, two different classes of ideas have been proposed. 

\begin{enumerate}
  \item Regularizers to learn domain-invariant mechanisms across domains.
  \item Data augmentation/manipulation techniques that reduce non-iidness in the data.
\end{enumerate}


For the first class of methods i.e., the design of novel regularizers for ERM,  multiple approaches have been proposed to learn the invariant mechanism \citep{rame2022fishr} from different perspectives. Despite the tremendous effort in developing novel regularizers, \citet{gulrajani2020search} shows that the second stream of methods e.g., adding a simple data augmentation (DA) to ERM, surprisingly outperforms or remains competitive compared to many of the aforementioned more complicated methods in the first class. Unlike the huge potential of the second approach, the majority of research ideas have been devoted to developing novel learning algorithms and few papers have explored the potential of novel data augmentation techniques \citep{zhang2017mixup}.


The above two directions follow the same objective of \emph{learning invariant mechanisms} which is usually achieved by removing the variation across training domains. However, for the first stream, this is not evident among these proposed regularizers, which of them can capture all the invariances.  As a matter of fact, we argue such a regularizer does not exist. This is because for each dataset, a wide range of shifts (correlation shift, diversity shift, label shift, etc.) across domains exist, and a rigid data-independent regularizer cannot remove all types of spurious correlations and shifts. Moreover, introducing these sub-optimal regularizers can make the optimization even more challenging by introducing excessive risk \citep{sener2022domain}, additional hyperparameters, and computational bottlenecks in ERM. The results presented in the DomainBed benchmark \citep{gulrajani2020search} validate this claim. Even though many different regularizers have been proposed, none of them achieve consistent superiority across all datasets. 





\begin{figure}[t]
\centering
\includegraphics[width=0.75\columnwidth]{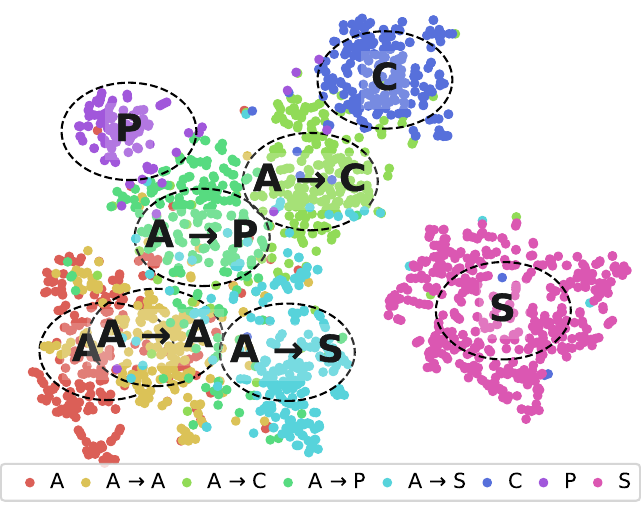} 
\caption{The $t$-SNE plot of features extracted from the original PACS dataset and generated images by the LDM from A domain. This figure shows that CDGA can fill the gap between domains. Check Section \ref{sec:domainquant} for details.}
\label{fig:tsne}
\end{figure}

In this paper, we explore the second direction, i.e., data augmentation, which is theoretically motivated by Vicinal Risk Minimization (VRM) \citep{vapnik1999nature,chapelle2000vicinal} principle. VRM reduces the estimation error of true data distribution in ERM by adding additional synthetic examples from the vicinity distribution around each observation \textbf{within each domain} in the data. However, in the DG setting, the estimation error of ERM is mainly caused by the distribution shift \textbf{between domains} which cannot be fully addressed by simple data augmentation techniques. We believe that with the help of recent off-the-shelf, pretrained generative models, e.g., latent diffusion model (LDM) \citep{rombach2022high}, we can implement the VRM principle on the domain level and subsequently achieve our goal of removing variability across domains. As illustrated in Figure \ref{fig:illustration}, we propose \emph{Cross Domain Generative Augmentation} (CDGA) which generates synthetic samples that fill the gap between all domain pairs for each class such that the domain shift is reduced. We illustrate the ability of CDGA to fill the gap across domains in Figure \ref{fig:tsne}, where it is shown that the generated images from domain A, i.e., A $\rightarrow$ A, A $\rightarrow$ P, A $\rightarrow $ C, and A $\rightarrow $ S fill the gap between the original domains, i.e., P, A, C, and S. We utilize two methods to implement CDGA: (1) \textbf{Prompt-Guided CDGA (CDGA-PG):} given an image from a domain and class, we grab the text descriptions of other domains as prompts to augment the current image to the images of other domains in the same class; (2) \textbf{Image-Guided CDGA (CDGA-IG):} in the case when domain text descriptions are not available, we utilize image mixer technique \citep{Image_Mixer} to mix each image with images from other domains that have the same class. Then, we utilize both generated synthetic and original datasets for training. We show such deep generative data augmentation techniques significantly improve the OOD accuracy of ERM and achieve SOTA performance on the DomainBed benchmark. We summarize our main contributions as follows:

\begin{itemize}

\item We propose CDGA that leverages the LDM \citep{rombach2022high} to generate synthetic samples to fill the gap between domains and reduce non-iidness in the training domain.
\item We show that CDGA along with vanilla ERM outperforms current SOTA DG algorithms through the DomainBed benchmark.
\item We generate more than 5 million synthetic images and perform extensive ablation studies to analyze the reasons why CDGA is effective, including data scaling law, domain shift quantification, adversarial robustness, mitigating class imbalance problem, and loss landscape analysis.
\end{itemize}

\begin{figure*}[ht!]
\centering
\includegraphics[width=0.99\textwidth]{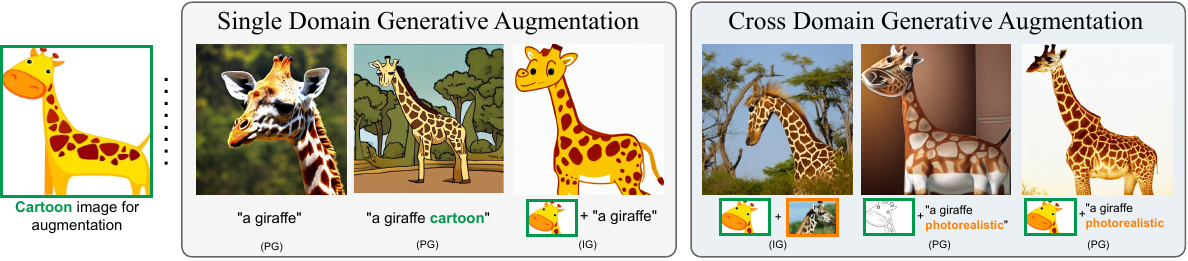} 
\caption{We compare different CDGA and SDGA techniques. The left column shows the original train image in the cartoon domain intended for generative augmentation. The middle and right columns show different variations of SDGA and CDGA respectively with their prompts.}
\label{fig2}
\end{figure*}

\section{Problem Settings and Related Work}
\label{sec:Problem Setting and Related Work}

We denote our prediction model as $f$ and its parameters as $\thetav$. In DG, the goal is to learn a shared model from $n$ source (train) domains (environments) $\{\Ec_1, \dots, \Ec_n\}$, to generalize to an unseen target domain $\Tc$. For a given domain $\Ec$, the classification loss is:
\begin{align}\label{eq:1}
\mathcal{L}_{\Ec}(\thetav) = \Eb_{(\xv, \yv)\sim \Ec} [\ell(f(\xv;\thetav), \yv)],
\end{align}
where each $\xv$ and $\yv$ are the input data point and its corresponding label and $\ell(f(\xv;\thetav), \yv)$ is the cross entropy loss between $f(\xv;\thetav)$ and $\yv$. 

 
 

\paragraph{The problem of domain shift.} The standard baseline for training deep learning models is Empirical Risk Minimization (ERM) \cite{vapnik1999overview}, which minimizes the average of losses over the entire training domains i.e., 
\begin{align}\label{eq:2}
\min_\thetav  \sum_i \frac{1}{| \Ec_i |} \sum_{k=1}^ {|\Ec_i|} \ell(f(x_{k}^{i};\thetav), y_{k}^{i})
\end{align} where $| \Ec_i |$ is the number of data points in domain $i$. $x_{k}^{i}$ is the $k$-th data point in domain $i$ and $y_{k}^{i}$ is its associated label. However, in the case of domain shift between domains, the pointwise kernel estimation of true data distribution proposed by ERM in Eq. \ref{eq:2},  becomes less accurate which results in a lack of OOD generalization of ERM. To improve the true data distribution estimation in ERM, \citet{vapnik1999nature} and \citet{chapelle2000vicinal} introduced VRM which proposes to replace the point-wise kernel estimates with a density estimation in the vicinity distribution around each observation within each domain. In practice, this can be implemented by data augmentation which adds additional synthetic examples from the vicinity distribution around each observation within each domain.

\noindent \textbf{Domain generalization}. Some of the recent research directions to improve the OOD performance of ERM are
robust optimization  \citep{sagawa2019distributionally,hu2018does}, invariant representation learning on feature level \citep{sun2016deep, ganin2016domain, li2018domain, tzeng2014deep}, classifier head \citep{arjovsky2019invariant}, loss \citep{krueger2021out} gradient/Hessian \citep{parascandolo2020learning, shahtalebi2021sand, koyama2020out, shi2021gradient, hemati2023understanding} and data augmentation-based algorithms \citep{gulrajani2020search,somavarapu2020frustratingly,zhou2021domain,carlucci2019domain}. With regards to data augmentation-based techniques, \citet{gulrajani2020search} employed classic data augmentation on top of ERM and showed under DomainBed evaluation protocol this strategy achieves better results compared to many competitors. \citet{ilse2021selecting} proposed Select Data Augmentation which starts with a list of transformations and selects the one that destroys the validation accuracy the most. In Mixup \citep{zhang2017mixup}, given two data points from two classes (not necessarily two domains), a new mixed data which is a linear convex combination of the two images, and their associated soft labels are generated.  \citet{zhou2021domain} proposed MixStyle where they mix per sample feature statistics (mean and variance) of training data across domains.

\noindent\textbf{Denoising diffusion models and their applications.} Recent efforts on diffusion-based generative models  \citep{ho2020denoising,song2020denoising,rombach2022high,zhang2023adding}  have shown these models can achieve SOTA image quality with reasonable sampling time. Furthermore, multiple works like Unclip \citep{ramesh2022hierarchical} have combined Foundation models like Contrastive Language-Image Pre-training like CLIP \citep{radford2021learning} with stable diffusion. This has added new generative functionalities including image-to-image, text-to-image, image variation, and image mixer to the diffusion-based models. Recently application of diffusion models in the representation learning problem has been explored. For example, \citet{tian2023stablerep} proposed StableRep in a setup similar to SimCLR \citep{chen2020simple} and showed synthetic images from stable diffusion models can improve self-supervised learning.

\section{Our Method: Cross Domain Generative Augmentation} 
\label{sec:Method}

\noindent \textbf{Theoretical Motivation}: First note that we can rewrite ERM \citep{vapnik1999nature} loss in Eq.~\ref{eq:1} as
\begin{align}\label{eq:3}
\min_\thetav \sum_i \frac{1}{| \Ec_i |}   \sum_k^{|\Ec_i |} \int \ell(f(x;\thetav), y) \delta_{x_{k}^{i}}(x) \delta_{y_{k}^{i}}(y) dx dy,
\end{align} where the data distribution is replaced by a set of delta functions on each data point i.e., $\delta_{x_{k}^{i}}(x)$ for $k$-th data point in domain $i$. On the other hand, the VRM loss function is written as 
\begin{align}\label{eq:4}
\min_\thetav \sum_i \frac{1}{|\Ec_i |}   \sum_k^{|\Ec_i |} \int \ell(f(x;\thetav), y) P_{x_{k}^i}^{i} (x) \delta_{y_{k}^{i}}(y) dx dy,
\end{align}  where $P_{x_{k}^i}^{i} (x)$ is the density estimation in the vicinity distribution around data point $k$ within domain $i$ which replaces the point-wise kernel estimate of ERM i.e., $\delta_{x_{k}^{i}}(x)$. In practice, VRM is implemented by employing data augmentation along with ERM. Although VRM can improve OOD performance, we argue such a scheme still cannot fully remove the shift between domains. This is because the estimation error of true data distribution by ERM is mainly caused by the distribution shift \textbf{between domains} which cannot be fully addressed by simple data augmentation techniques within each domain. More specifically, classic augmentation only accepts one data point $\texttt{img}_{k}^{i}$ from domain $i$ and returns $\widehat{\texttt{img}}_{k}^{i} = g(\texttt{img}_{k}^{i})$ where $g(\cdot)$ is a simple transformation and $\widehat{\texttt{img}}_{k}^{i}$ is the augmented data point.

\noindent \textbf{CDGA}: To tackle the domain shift between all domain pairs, we propose the following loss function
\begin{align}\label{eq:5}
\min_\thetav \sum_i \sum_j \frac{1}{| \Ec_i |}   \sum_k^{| \Ec_i |} \int \ell(f(x;\thetav), y_{k}^{i}) P_{x_{k}^{i}}^{i,j}(x) \delta_{y_{k}^{i}}(y) dx dy,
\end{align} where $i$ and $j$ are domain counters and $k$ is the data point counter, $ P_{x_{k}^{i}}^{i,j}(x)$ represents the vicinity of domain pair $i$ and $j$ for $k$-th data point in domain $i$. Considering that we do not have access to $ P_{x_{k}^{i}}^{i,j}(x) $, the minimization of the above loss function is not tractable in practice. However, similar to VRM, a simple practical implementation of the above loss function is to generate samples drawn from the vicinity of domain pairs. In other words, our goal is to generate synthetic images in a way that fills the gap between all possible domain pairs and reduces the non-iidness. Thanks to image manipulation of diffusion models, such sampling has become feasible. To this end, in CDGA, the simple transformation $g(\cdot)$ is replaced with the LDM, denoted by $\Mc(\cdot)$ which takes two arguments, one is a data point in a domain and the second argument is a guidance attribute in another domain from the same class i.e.,
\begin{align}\label{eq:6}
\widehat{\texttt{img}}_{k}^{i,j} = \Mc(\texttt{img}_{k}^{i},\texttt{guide}^{j}),
\end{align} where $\widehat{\texttt{img}}_{k}^{i,j}$ is a synthetic image that interpolates domains $i$ and $j$, generated from $k$-th sample in domain $i$ and $\texttt{guide}^{j}$ is the guidance towards another domain $j$ within the same class. In other words, $\widehat{\texttt{img}}_{k}^{i,j}$ is drawn from $P_{x_{k}}^{i,j} (x) $. In CDGA, each data point in domain $i$ is transformed to all $n$ domains (including its own domain which increases the number of samples for the domain $\Ec_i$ from $| \Ec_i |$ to  ($b \times n+1) \times | \Ec_i |$ where $n$ is the number of training domains, $| \Ec_i |$ is the number of data points in domain $i$, and $b$ is the generation batch size. In addition to CDGA, we propose CDGA$^*$ where we assume we may have access to a guidance attribute of the target domain. In this case, the size of the domain $\Ec_i$ increases from $| \Ec_i |$ to $(b \times n+2) \times | \Ec_i |$. CDGA guidance can be a real image or text prompt as shown in Figure \ref{fig2}.




\subsection{CDGA with Prompt Guidance (CDGA-PG)}

In CDGA-PG, given the $k$-th image in domain $i$, i.e., $\texttt{img}^{k}_{i}$, the guidance attribute $\texttt{guide}^{j}$ is a domain description text prompt that represents the same class in domain $j$, e.g. ``a cat in painting style.'' Then, having the image and the prompt guidance we use the LDM to generate $b$ synthetic images which we expect to interpolate domains $i$ and $j$ for the same class. For each image in domain $i$, we perform this interpolation for all the training domains $j$, $\forall j \in \{1, ..., n \}$. We also consider the scenario where we can utilize the target domain description, i.e., $\texttt{guide}^{target}$ as the guidance. This method named CDGA$^*$-PG is the same as CDGA-PG with the difference that we also have access to the target domain description. This provides us the opportunity to directly interpolate between training domains and the target domain.




\subsection{CDGA with Image Guidance (CDGA-IG)}

For scenarios where a text prompt description of domains is not available, e.g., domains are hospitals,  CDGA-PG will not be applicable. To address this challenge, we propose CDGA-IG where the guidance is an image from domain $j$ instead of a text description. More precisely, in CDGA-IG we attempt to mix two images from two different domains which is also known as the image mixer in the literature. 


\section{Single Domain Generative Augmentation (SDGA)}
We also explore SDGA method, where unlike CDGA, the image from domain $i$ is augmented only from the guidance of the same domain $\Ec_i$, i.e., 
\begin{align}\label{eq:7}
\widehat{\texttt{img}}_{k}^{i} = \Mc (\texttt{guide}^{i}),
\end{align}
where guidance can either be an image (IG) or a prompt (PG).

\noindent \textbf{SDGA-PG}: In this technique, for each image in domain $i$, i.e., $\texttt{img}_{k}^{i}$ we create prompt guidance $\texttt{guide}^i$ that can contain label and/or domain information from domain $i$ and feed $\texttt{guide}^i$ to the LDM. 

\noindent\textbf{SDGA-IG}: Here, for each image from domain $i$, i.e., $\texttt{img}_{k}^{i}$ we construct a guidance that contains both $\texttt{img}_{k}^{i}$ and label information.



\section{Experiments}

\begin{figure}[t]
\centering
\includegraphics[width=0.99\columnwidth]{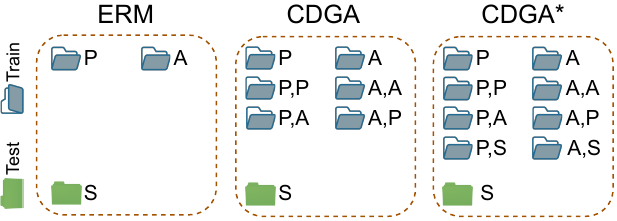}
\caption{Illustration of the implementation structure of ERM, CDGA, and CDGA$^*$ on PACS dataset when using P and A domains as training and S as target domain.}
\label{fig:implementation}
\vspace{-1em}
\end{figure}

\noindent  \textbf{Datasets}: We evaluate our method on multiple datasets including VLCS \citep{fang2013unbiased}, PACS \citep{li2017deeper},  OfficeHome \citep{venkateswara2017deep}, and DomainNet \citep{peng2019moment}.

\noindent \textbf{Models}: We use the pretrained, version 1.4 of stable diffusion \citep{rombach2022high} without finetuning as our base LDM.  For the implementation of CDGA-IG, we use the image mixer that has been fine-tuned by Justin Pinkney at Lambda Labs \citep{Image_Mixer} to accept CLIP image embeddings. For image generation, we do not tune any hyperparameters (e.g., strength, steps, etc) and all the parameters are set to their default values of the \citep{rombach2022high} repository.

\noindent \textbf{Prompts}: For CDGA-PG, we use both the classes of the images and the domain description in the text prompts as guidance. The complete list of the prompts used for each domain in each dataset is in appendix \ref{subsec:prompts}.

\noindent \textbf{Implementation method}: For implementing CDGA, we use offline augmentation where we first generate images between each pair of training domains, and then start the training process. The folder structure of our implementation for the PACS dataset when using P and A domains as train domains and S domain for test domain is illustrated in Figure \ref{fig:implementation}. For all the methods, we set generation batch size $b=1$ unless stated otherwise. 

\noindent \textbf{Hardware}: We use two clusters of four V100 NVIDIA GPUs for generation and benchmarks.

\noindent In all the tables, we format \textbf{first}, \underline{second} results.

\label{sec:Results}
\subsection{DomainBed Benchmark} \label{sec:domainbed}

The DomainBed benchmark \citep{gulrajani2020search} has become a common way to evaluate domain generalization algorithms in a fair and standard manner. Given a dataset, this evaluation scheme compares DG algorithms through three different model selection techniques over 20 choices of hyperparameters and 3 trials. To show the effectiveness of CDGA, we evaluate it using the DomainBed benchmark in Tables \ref{table:Table1}- \ref{table:Table4}. Except for the DomainNet dataset, our setup is exactly the same as the DomainBed benchmark for all datasets. For the DomainNet dataset, our CDGA method generated more than 4 million images which made running the ERM for all possible grids of 20 choices of hyperparameters and 3 trials over CDGA data very expensive. To overcome this computational bottleneck, for DomainNet we ran the same leave-one-domain-out experiment for ERM with 1 choice of hyperparameter (default hyperparameters) and 1 trial. To save space, here \textbf{we only report the five top-performing} algorithms for each model selection and the full results are available in the appendix. Looking at Tables \ref{table:Table1}- \ref{table:Table4}, CDGA achieves SOTA in all datasets and model selection techniques. Here for the PACS, OfficeHome, and DomainNet, we employed the prompt guidance while for the VLCS, the image guidance (image mixer) was used. The code implementation for deploying CDGA-generated data in the DomainBed scheme is presented in \Cref{subsec:Code}.


\begin{table}[h]%
    \caption{DomainBed benchmark for \textbf{training-domain validation set} model selection method. }
    \centering
    {
        \resizebox{\columnwidth}{!}{\begin{tabular}{l|cccc}
            \toprule
            Algorithm                                  & \textbf{PACS}                                 & \textbf{OfficeHome}  & \textbf{DomainNet}                                                      & \textbf{Avg}               \\
            \midrule
            ERM                                                               & 85.5 \scriptsize{$\pm$ 0.2}                   & 66.5 \scriptsize{$\pm$ 0.3}
            & 40.9 \scriptsize{$\pm$ 1.8}                                   & 64.3                               \\
            CORAL                                                     & 86.2 \scriptsize{$\pm$ 0.3}       & \underline{68.7} \scriptsize{$\pm$ 0.3}                      & 41.5 \scriptsize{$\pm$ 0.1}                   & 65.5           \\
            SagNet                                                            & 86.3 \scriptsize{$\pm$ 0.2}          & 68.1 \scriptsize{$\pm$ 0.1}                      & \textcolor{gray}{40.3} \scriptsize{$\pm$ 0.1} & 64.9            \\
            Fish                              & 85.5 \scriptsize{$\pm$ 0.3}             &68.6 \scriptsize{$\pm$ 0.4}         &42.7    \scriptsize{$\pm$ 0.2}            &  65.6                  \\
        
            Fishr                                              & 85.5 \scriptsize{$\pm$ 0.4}           & 67.8 \scriptsize{$\pm$ 0.1}                                  & 41.7 \scriptsize{$\pm$ 0.0}       & 65.0                             
                          \\ 
             HGP                                       & 84.7   \scriptsize{$\pm$ 0.0}              & 68.2   \scriptsize{$\pm$ 0.0}              &41.1   \scriptsize{$\pm$ 0.0}     & 64.7                     \\

                \midrule

         CDGA-PG                   & \underline{88.5}\scriptsize{ $\pm$ 0.5}    &  68.2 \scriptsize{$\pm$  0.6    }                                  & \underline{43.1} \scriptsize{$\pm$0.0}       & \underline{66.6}   \\

      CDGA-PG$^*$                   & \textbf{89.5}\scriptsize{ $\pm$ 0.3}            & \textbf{70.8} \scriptsize{$\pm$ 0.6 }                                 & \textbf{44.8} \scriptsize{$\pm$0.0}       & \textbf{68.4}  \\

     \bottomrule
        \end{tabular}
    }}
    \label{table:Table1}%
\end{table}

\begin{table}[ht]%
    \caption{DomainBed benchmark for \textbf{leave-one-domain-out cross-validation} model selection.}%
    \centering
    {
         \resizebox{\columnwidth}{!}{\begin{tabular}{l|ccccc}
            \toprule
            Algorithm & \textbf{PACS}                                 & \textbf{OfficeHome}   & \textbf{DomainNet}                                                     & \textbf{Avg}               \\
            \midrule
            ERM                           & 83.0  \scriptsize{$\pm$ 0.7}            & 65.7 
  \scriptsize{$\pm$ 0.5}                 & 40.6 \scriptsize{$\pm$ 0.2}         & 63.1                    \\
          
            CORAL                                   & \textcolor{gray}{82.6} \scriptsize{ $\pm$ 0.5}       & 68.5 \scriptsize{$\pm$ 0.2}    & 41.1 \scriptsize{ $\pm$ 0.1}                 & 64.1       \\
            SagNet                                      &\textcolor{gray}{82.3} \scriptsize{ $\pm$ 0.1}                 & 67.6 \scriptsize{ $\pm$ 0.3}     & 40.2  \scriptsize{ $\pm$ 0.2}  & 63.4                              \\

        MLDG                                      & \textcolor{gray}{82.9}   \scriptsize{ $\pm$ 1.7}         & 66.1 \scriptsize{ $\pm$ 0.5}     &41.0  \scriptsize{ $\pm$ 0.2}        & 63.3   
\\
         HGP                                               & 82.2   \scriptsize{ $\pm$ 0.0}          & 67.5  \scriptsize{ $\pm$ 0.0}            &41.1   \scriptsize{$\pm$ 0.0}       &63.6                    \\
           Hutchinson                                      & 84.8  \scriptsize{ $\pm$ 0.0}             & 68.5    \scriptsize{ $\pm$ 0.0}           &41.4    \scriptsize{ $\pm$ 0.0}     & 64.9  \\
             \midrule

             CDGA-PG                                                               &  \underline{86.8} \scriptsize{ $\pm$ 0.4}                   &  \underline{68.7}   \scriptsize{ $\pm$ 0.4}             & \underline{43.1} \scriptsize{$\pm$0.0} & \underline{66.2}                   \\
            
             CDGA-PG$^*$                                                   &  \textbf{88.4}   \scriptsize{$\pm$ 0.5}                & \textbf{70.2} \scriptsize{ $\pm$ 0.4}          & \textbf{44.8} \scriptsize{$\pm$0.0}          & \textbf{67.8}                    \\
           \bottomrule
        \end{tabular}}
    }
    \label{table:Table2}%
\end{table}

\begin{table}[ht]%
    \caption{DomainBed benchmark \textbf{test-domain validation set (oracle)}model selection method.}%
    \centering
    {
         \resizebox{\columnwidth}{!}{\begin{tabular}{l|cccc}
            \toprule
            Algorithm  & \textbf{PACS}     &\textbf{OfficeHome}  & \textbf{DomainNet}                                                      & \textbf{Avg}               \\
            \midrule
            ERM                                        & 86.7 \scriptsize{$\pm$ 0.3}                   & 66.4 \scriptsize{$\pm$ 0.5}                               & 41.3 \scriptsize{$\pm$ 0.1}                   & 64.8             \\
            
            Mixup                                                                   & 86.8 \scriptsize{$\pm$ 0.3}                   & 68.0 \scriptsize{$\pm$ 0.2}                          & \textcolor{gray}{39.6} \scriptsize{$\pm$ 0.1}     & 64.8         \\
            MLDG                                                            & 86.8 \scriptsize{$\pm$ 0.4}                   & 66.6 \scriptsize{$\pm$ 0.3}                     & 41.6 \scriptsize{$\pm$ 0.1}                   & 65.0             \\
            CORAL                                        & 87.1 \scriptsize{$\pm$ 0.5}                   & 68.4 \scriptsize{$\pm$ 0.2}          & 41.8 \scriptsize{$\pm$ 0.1}       & 65.8         \\
            SagNet                                          & \textcolor{gray}{86.4} \scriptsize{$\pm$ 0.4} & 67.5 \scriptsize{$\pm$ 0.2}                 & \textcolor{gray}{40.8} \scriptsize{$\pm$ 0.2} & 64.9                            \\
            
            Fish                                     & 85.8 \scriptsize{$\pm$ 0.6}                   & 66.0 \scriptsize{$\pm$ 2.9}        & \underline{43.4} \scriptsize{$\pm$ 0.3}          &  65.1               \\
        
            Fishr                                & 86.9 \scriptsize{$\pm$ 0.2} & 68.2 \scriptsize{$\pm$ 0.2} & 41.8 \scriptsize{$\pm$ 0.2}          & 65.6                            
                          \\ 
             Hutchinson                        & \textcolor{gray}{86.3}  \scriptsize{$\pm$  0.0}            & 68.4 \scriptsize{$\pm$  0.0}    &41.9 \scriptsize{$\pm$  0.0}            & 65.5                \\
                 \midrule
     CDGA-PG         &\underline{89.6 }\scriptsize{$\pm$  0.3   }            &   \underline{68.8}  \scriptsize{$\pm$ 0.3}                                  & 43.1 \scriptsize{$\pm$0.0}       & \underline{67.2}  \\ 

      CDGA-PG$^*$                  &   \textbf{90.4} \scriptsize{$\pm$ 0.3}  & \textbf{70.2}  \scriptsize{$\pm$ 0.2}                                  & \textbf{44.8} \scriptsize{$\pm$0.0}       & \textbf{68.5}  \\
     \bottomrule
        \end{tabular}
    }}
    \label{table:Table3}%
\end{table}

\begin{table}[ht]
\caption{DomainBed benchmark on \textbf{VLCS} dataset.}

  \centering
  \scalebox{0.8}{\begin{tabular}{lccc}
    \toprule
     Method & \begin{tabular}{@{}c@{}}Training \\ domain\end{tabular} & \begin{tabular}{@{}c@{}}Leave-one \\ -domain-out\end{tabular}  &Oracle    \\
    \midrule
         ERM &  77.5  \scriptsize{$\pm$ 0.4}  &  77.2 \scriptsize{$\pm$ 0.4}  &   77.6 \scriptsize{$\pm$ 0.3}  \\ 

       CORAL  & \underline{78.8}   \scriptsize{$\pm$ 0.6}  &  \underline{78.7}  \scriptsize{$\pm$ 0.4}  &  77.7  \scriptsize{$\pm$ 0.2}  \\
              SagNet  & 77.8   \scriptsize{$\pm$ 0.5}  &  77.5  \scriptsize{$\pm$ 0.3}  &   77.6  \scriptsize{$\pm$ 0.1}  \\ 
       Fishr  & 77.8   \scriptsize{$\pm$ 0.1}  &  78.2  \scriptsize{$\pm$ 0.0}  &   \underline{78.2}  \scriptsize{$\pm$ 0.2}  \\ 
       HGP  & 77.6   \scriptsize{$\pm$ 0.0}  &  76.7  \scriptsize{$\pm$ 0.0}  &   77.3  \scriptsize{$\pm$ 0.0}  \\
        Hutchinson  & \textcolor{gray}{76.8}   \scriptsize{$\pm$ 0.0}  &  \textbf{79.3}  \scriptsize{$\pm$ 0.0}  &   77.9  \scriptsize{$\pm$ 0.0}  \\

    \midrule

          CDGA-IG &\textbf{78.9}   \scriptsize{$\pm$ 0.3} & 77.9  \scriptsize{$\pm$ 0.5} & \textbf{79.5}   \scriptsize{$\pm$ 0.1}\\
    \bottomrule
  \end{tabular}}

 \label{table:Table4}
\end{table}

\subsection{CDGA vs SDGA}

To compare and evaluate variations of CDGA and SDGA, we compare them on the PACS dataset using the DomainBed benchmark with twenty different hyperparameters and one trial.  The results of this experiment are presented in Table \ref{table:Table6}. Due to the domain interpolation ability in CDGA, it consistently outperforms all variations of SDGA. Among SDGA-based techniques, SDGA-IG that only uses the class label information outperforms other SDGA-based methods.



\begin{table}[ht]
\caption{OOD accuracy of models trained with variations of CDGA and SDGA on PACS dataset using Domainbed benchmark.}
  \centering
  \resizebox{\columnwidth}{!}{\begin{tabular}{lccc}
    \toprule
     Method & \begin{tabular}{@{}c@{}}Training \\ domain\end{tabular} & \begin{tabular}{@{}c@{}}Leave-one \\ -domain-out\end{tabular}  &Oracle    \\
    \midrule
        ERM &  85.5   \scriptsize{$\pm$ 0.2}  &  83.0   \scriptsize{$\pm$ 0.7}  &   86.7   \scriptsize{$\pm$ 0.3}  \\ 
          \midrule
        SDGA-PG (label) &  86.1   \scriptsize{$\pm$ 0.5}  &  83.7   \scriptsize{$\pm$ 1.0}  &   87.3   \scriptsize{$\pm$ 0.5}  \\ 
       SDGA-PG (label $+$ domain)  & 85.9   \scriptsize{$\pm$ 1.0}  & 84.6   \scriptsize{$\pm$ 0.8}  &  87.5   \scriptsize{$\pm$ 1.1}   \\ 
        SDGA-IG (label) & 87.5   \scriptsize{$\pm$ 0.6}  & 86.5   \scriptsize{$\pm$ 1.1}   & 89.5    \scriptsize{$\pm$ 0.3} \\ 
          \midrule
        CDGA-PG (canny edge) & 86.8   \scriptsize{$\pm$ 0.9}  & 79.2   \scriptsize{$\pm$ 3.6}   & 87.8    \scriptsize{$\pm$ 0.3} \\ 
        CDGA-PG &  \underline{88.5}   \scriptsize{$\pm$ 0.5}    &  \underline{86.8}   \scriptsize{$\pm$ 0.4}  &  \underline{89.6}   \scriptsize{$\pm$ 0.3}  \\ 
        \midrule
       CDGA$^*$-PG & \textbf{89.5}   \scriptsize{$\pm$ 0.3} & \textbf{88.4}   \scriptsize{$\pm$ 0.5}   & \textbf{90.4}   \scriptsize{$\pm$ 0.3}  \\
    \bottomrule
  \end{tabular}}

 \label{table:Table6}
\end{table}


\subsection{Scaling Law of Data Size}
In this experiment, we want to measure the effect of generation batch size $b$ which determines the final augmented dataset size on the OOD generalization of models trained with CDGA. To this end, on the PACS dataset, using DomainBed benchmark with 20 choices of hyperparameters and 1 trial, we apply CDGA-PG with $b$ equal to 1, 2, 3, and 4. We conduct this experiment for two models:  ResNet-18 model pretrained on ImageNet and  ResNet-18 model with random initialization. As we see from Table \ref{table:Table7}, for both pretrained and random initializations, increasing $b$, i.e., increasing the data size, further improves OOD generalization. However, for the pretrained model, the performance gets saturated which can be due to the capacity of the model.

\begin{table}[ht]
\caption{Effect of generation batch size $b$ on CDGA for PACS dataset with different initialization.}
  \centering
\begin{tabular}{lcccc}
    \toprule
    \small
    Initialization &$b$ & \begin{tabular}{@{}c@{}}Training \\ domain\end{tabular} & \begin{tabular}{@{}c@{}}Leave-one \\ -domain-out\end{tabular}  &Oracle    \\
    \midrule
    \multirow{4}{*}{Random}  &1 &66.1     & 61.6   & 65.0    \\ 
    &2   & 70.2   & 69.2    &  70.3 \\ 
    &3  &   \underline{71.8}   &  \underline{71.8}  & \underline{74.5} \\ 
    &4  & \textbf{73.9} &\textbf{73.2} & \textbf{74.6}  \\
    \midrule
    \multirow{4}{*}{Pre-trained}
    &1 & 87.0  & 86.4  &  \underline{89.0}   \\ 
    &2   &  \underline{87.4}  & 86.2    & \underline{89.0} \\ 
    &3  & \textbf{88.4}     & \textbf{89.1}   & 88.9 \\ 
    &4  & \textbf{88.4}  & \underline{88.2}    &\textbf{89.2}  \\
    \bottomrule
  \end{tabular}

 \label{table:Table7}
\end{table}

\subsection{Mitigating Class Imbalance}
CDGA can also be utilized to mitigate the class imbalance problem in datasets where the number of instances in each class of each domain is not equal. In such scenarios, one can use a different $b$ for each class of the data such that after generating samples, the number of instances in each class of generated domains becomes equal. We test the effectiveness of CDGA method in balancing the OfficeHome dataset (which is highly imbalanced) through the DomainBed benchmark. More specifically, for every class $c$ and domain $\Ec_j$, we find the number of samples $n(\Ec_j, c)$ and then we find $m = \max_{c, j} \ n(\Ec_j, c)$ which is 100 for OfficeHome. Then for every domain $\Ec_j$ and class $c$ we set $b=\frac{m}{n(\Ec_j, c)}$ which leads to larger batch size for domains and classes with fewer data points and subsequently balances the dataset. The results of this experiment are presented in Table \ref{table:Table5}. Clearly, by choosing $b$ in a way that the dataset is more balanced, the OOD generalization has been further improved.
\begin{table}[ht]
\caption{OOD accuracy of models with and without balanced generation in OfficeHome dataset .}
  \resizebox{\columnwidth}{!}{\begin{tabular}{lccc}
    \toprule
     Method & \begin{tabular}{@{}c@{}}Training \\ domain\end{tabular} & \begin{tabular}{@{}c@{}}Leave-one \\ -domain-out\end{tabular}  &Oracle    \\
    \midrule
        ERM  & 66.5 \scriptsize{$\pm$ 0.3}  &  65.7 \scriptsize{$\pm$ 0.5}         & 66.4 \scriptsize{$\pm$ 0.5}    \\ 
        \midrule
        CDGA-PG ($b=1$)  &\underline{68.2} \scriptsize{$\pm$ 0.6}  &  \underline{68.7} \scriptsize{$\pm$ 0.4}  & \underline{68.6} \scriptsize{$\pm$ 0.3}  \\ 
       CDGA-PG ($b=\frac{m}{n(\Ec_j, c)}$)  & \textbf{69.9} \scriptsize{$\pm$ 0.2} & \textbf{69.7} \scriptsize{$\pm$ 0.4} &  \textbf{70.0} \scriptsize{$\pm$ 0.7} \\
    \bottomrule
  \end{tabular}}

 \label{table:Table5}
\end{table}




\subsection{Domain Shift Quantification}
\label{sec:domainquant}

To show the effectiveness of CDGA in mitigating domain shift and reducing non-iidness, we need tools to quantify the shift between domains. Here we apply a wide range of domain shift quantification techniques proposed in the literature on the PACS dataset. More precisely, we quantify the domain shift using t-SNE visualization of feature embeddings, near duplicate analysis \citep{oquab2023dinov2}, transferability \citep{zhang2021quantifying}, \citep{hemati2023understanding}, and diversity shift metrics \citep{ye2022ood}. 


\noindent \textbf{Domain shift visualization.} To visualize the domain shift for CDGA-based data, only for the class dog, given data from all domains P, A, C, and S, we consider synthetic images for A $\rightarrow  $ A, A $\rightarrow  $ P, A $\rightarrow  $ C, and A $\rightarrow $ S. Then, we feed the real and synthetic images to the pretrained CLIP ViT-B/32 image encoder \citep{radford2021learning} to extract features from images and use t-SNE to project them on two-dimensional space. The projected features are plotted in Figure \ref{fig:tsne}. Interestingly the cross-domain synthetic images interpolate different domains as we desired. For example, in Figure \ref{fig:tsne}, consider domains A (in red) and S (in pink), even though all images are in dog class, there is a significant distribution shift between their two-dimensional representations. However, A $\rightarrow $ S synthetic images perfectly fill the gap between A and S representations. The same observation consistently can be made for other domain pairs which imply that CDGA-based synthetic images are contributing to reducing non-iidness. See Figure~\ref{fig:tsne_full} in the appendix for the t-SNE plots of other classes.



\begin{figure}[t]
\centering
\includegraphics[width=0.99\columnwidth]{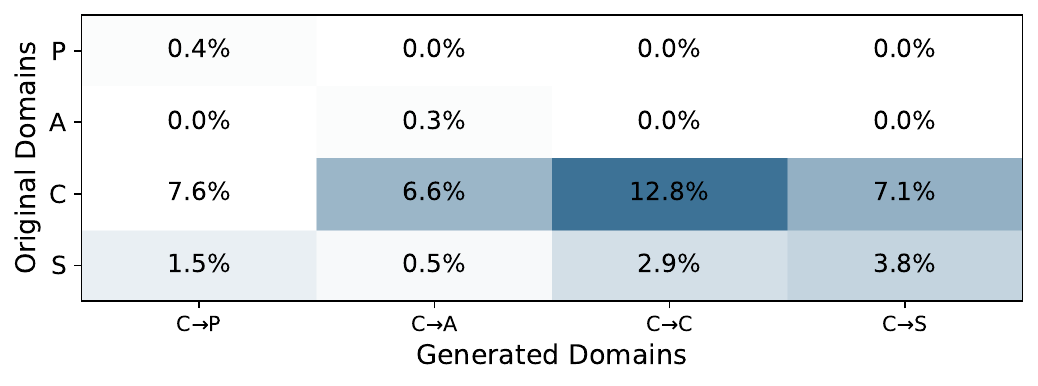} 
\caption{Heat map of the percentage of near-duplicates of each original domain in the generated domains. This table shows that using target-domain description results in more near-duplicate images.}
\label{fig:dup_heatmap}
\end{figure}

\begin{figure}[ht]
\centering
\includegraphics[width=0.85\columnwidth]{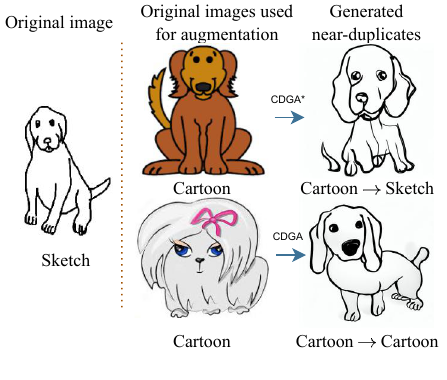} 
\caption{Examples of near-duplicates (right-most column) found for the dog image in Sketch domain (left-most column) that are generated using CDGA from the original images (middle column).}
\label{fig:near_dup_illus}
\end{figure}


\noindent \textbf{Near-duplicate Analysis.} We apply near-duplicate image detection to images generated using CDGA to quantify how much the generated images are similar to the original images for each domain. Following the self-supervised image retrieval technique in \citep{oquab2023dinov2}, we utilize the pretrained CLIP  ViT-B/32 image encoder \citep{radford2021learning} to extract embeddings and calculate the cosine similarity between each of the original images and the generated images; Then, for each original image, if there is at least one image in a generated domain that is above 0.95 cosine similarity we say the original domain has a near-duplicate. A summarized version of this experiment for the case that the generated images are from domain C is presented in Figure \ref{fig:dup_heatmap} and the full result can be found in Figure \ref{fig:dup_heatmap_full} in the appendix. In this figure, for each original domain, we report the percentage of the number of near-duplicates over the size of the original domain. Clearly, generating synthetic images that exist in the manifold between training domains enables us to have examples near-duplicate to the target domain. Figure \ref{fig:near_dup_illus} shows some of the near-duplicates found for a sample image in the S domain using this technique. For more examples see Figure \ref{fig:near_dup_illus_full} in the appendix.



\begin{figure}[t]
\centering
\includegraphics[width=0.85\columnwidth]{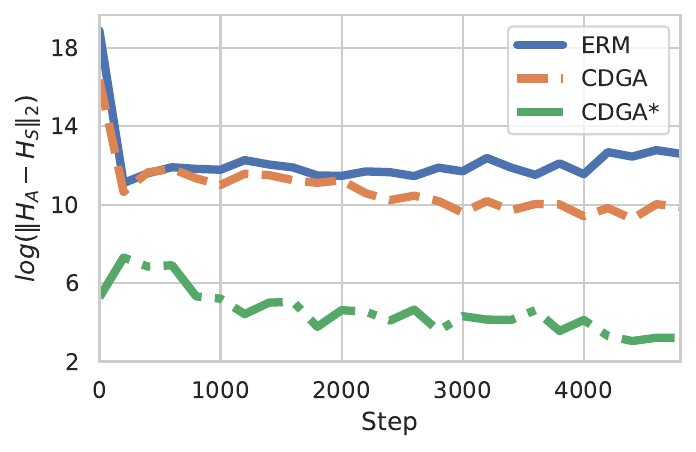}
\caption{Classifier head Hessian difference between domains P \& S during training for PACS and CDGA-based PACS datasets i.e., CDGA-PACS and CDGA$^*$-PACS where Domains A, P, and C are training domains and domain S is the target domain.}
\vspace{-1em}
\label{fig:Hessian_diff}
\end{figure}

\noindent \textbf{Transferability.} Transferability is another recent approach proposed by \citet{zhang2021quantifying} to quantify the domain shift. 
The transferability measure is an upper bound of the difference between the source and the target domain excess risks. Subsequently, \citet{hemati2023understanding} showed that an upper bound for transferability measure is $\frac{1}{2}\delta^2 \|\Hv_\Tc  - \Hv_\Sc\|_2 + o(\delta^2)$ where $\delta$ is a constant, $\Hv_\Tc$ and $\Hv_\Sc$ are target and source classifier head's Hessians. Following these findings, to quantify the dynamics of domain shift, we monitor classifier heads Hessian distances between all possible domain pairs through the training steps for ERM and CDGA, where we set domains P, A, and C as train (source) and domain S as target. Figure \ref{fig:Hessian_diff} shows the difference between the classifier head's Hessians given data from domains A and S during the steps. We see that CDGA and CDGA$^*$ lead to lower classifier head's Hessian difference and subsequently smaller transferability and domain shift compared to ERM. This pattern consistently exists for other domain pairs where the full results are presented in Figure \ref{fig:Hessian_diff_full} in the appendix.

\begin{figure}[t]
\centering
\includegraphics[width=0.85\columnwidth]{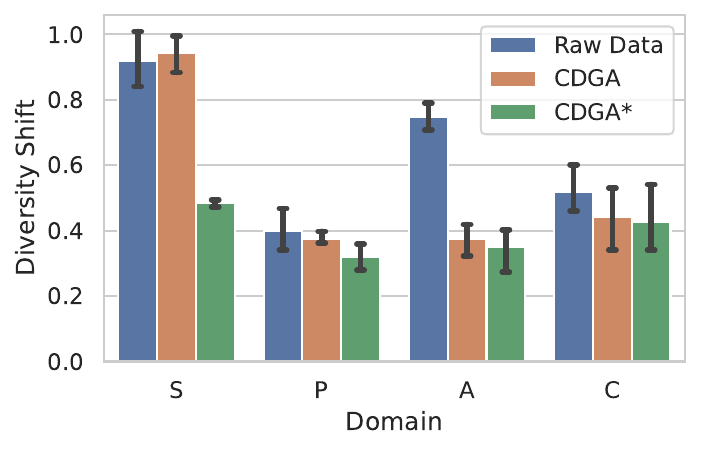}
\caption{Diversity shift as a measure of iid-ness of PACS (raw data) against CDGA, and CDGA$^*$ augmented datasets. Each column is the target domain and the rest of the domains are training domains. CDGA and CDGA$^*$ reduce the diversity shift.}
\label{fig:div_shift}
\vspace{-1em}
\end{figure}

\noindent \textbf{Diversity Shift.} \citet{ye2022ood} proposed a numerical method to measure diversity shift which is equivalent to total variation \citep{zhang2021quantifying} to quantify domain shift. Diversity shift is usually due to the novel domain-specific features in the data. We employ the proposed algorithm by \citet{ye2022ood} to quantify and compare diversity shift between training domains and the target domain in a leave-one domain out scheme for PACS real data, CDGA-PACA, and CDGA$^*$-PACS  datasets. Figure \ref{fig:div_shift} shows both CDGA and CDGA$^*$ reduce the diversity shift between training domains and the target domain.

\subsection{Adversarial Robustness of Models Trained With CDGA}

\begin{figure}[ht]
  \centering
  \medskip
  \begin{subfigure}[t]{0.49\linewidth}
    \centering\includegraphics[width=4.2cm]{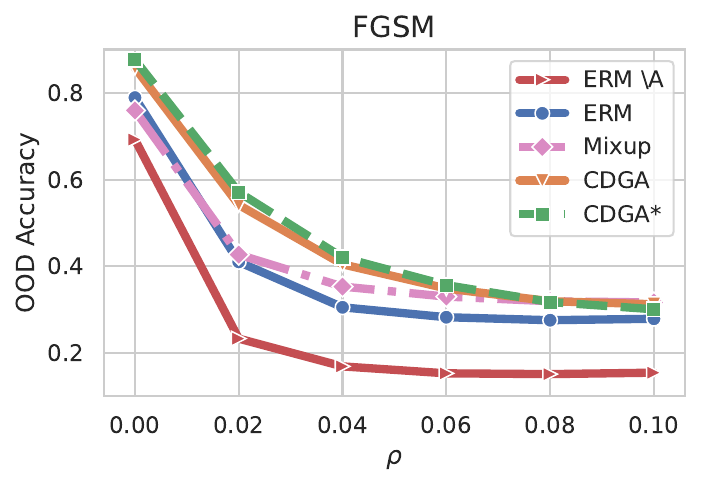}
    \caption{}\label{fig:fgsm}
  \end{subfigure}
  \begin{subfigure}[t]{0.49\linewidth}
    \centering\includegraphics[width=4.2cm]{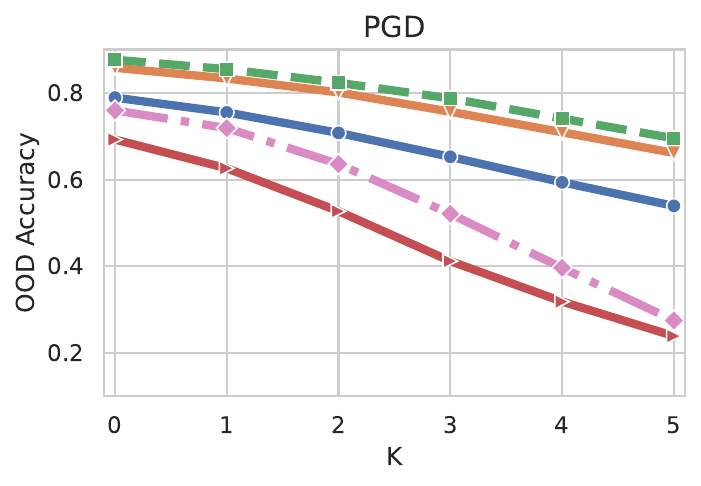}
    \caption{}\label{fig:pgd}
    \end{subfigure}
 \caption{OOD accuracy of models trained using CDGA, CDGA$^*$, classic augmentation, and MixUp after FGSM adversarial shift with varying perturbation size $\rho$ (Figure \ref{fig:fgsm}) and PGD adversarial shift with varying number of iterations $K$ (Figure \ref{fig:pgd}). Models trained with CDGA and CDGA$^*$ are more robust against adversarial shifts. ERM $\backslash$A means without data augmentation. }
 \vspace{-1.4em}
  \label{fig:advattack}%
\end{figure}

The robustness of deep learning models to adversarial attacks is a notion of their generalization ability \citep{shorten2019augsurvey}. Here, we study the robustness of models trained with CDGA and CDGA$^*$ against adversarial shifts to the data. To this end, we evaluate their performance on the target domain after shifting the target domain using the fast gradient sign method (FGSM) \citep{goodfellow2015fgsm}, and the more sophisticated projected gradient descent (PGD) adversarial attacks \citep{madry2018pgd}. We design an experiment using the PACS dataset where the training domains are P, A, and C  and the target domain is S. After training the same model using different data augmentation techniques including MixUp and the classic data augmentation, we perturb the target domain using FSGM and PGD and record the OOD accuracy for different strength of attacks. Results are presented in Figure \ref{fig:advattack}, where $\rho$ controls the FGSM perturbation size, and $K$ controls the number of iterations in PGD. Clearly, models trained with CDGA and CDGA$^*$ are more robust against adversarial shifts on the target domain compared with the classic data augmentation and Mixup.

\begin{figure}[t]
\centering
\includegraphics[width=0.85\columnwidth]{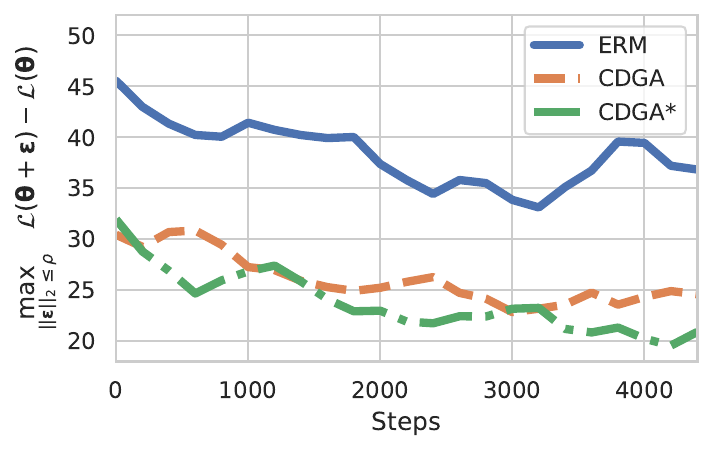}
\caption{Sharpness of loss landscape through training for ERM, CDGA, and CDGA$^*$ on PACS dataset.}
\label{fig:sharpness}
\vspace{-1em}
\end{figure}


\subsection{Loss Landscape Analysis: Does CDGA Guide us to Flatter Local Minimum?}
Recently, there has been a growing interest in finding the relation between loss landscape structure and generalization ability of deep neural networks \citep{keskar2016large,foret2020sharpness}. More precisely, according to the theorem stated by \citet{foret2020sharpness} generalization error of deep neural networks is upper bounded by the sharpness of their loss landscapes:
\begin{align}\label{eq:8}
\underset{\|\bm{\epsilon}\| _2 \le \rho}{\text{max}} \ \mathcal{L}(\thetav + \bm{\epsilon}) - \mathcal{L}(\thetav),
\end{align} where $\rho$ is the perturbation size.
In other words, neural networks with flatter local minima have less generalization error. Inspired by this finding, we raise the following question. Given better OOD generalization of CDGA and CDGA$^*$, do these methods lead to flatter local minima compared with ERM? To answer this question we calculate the sharpness using Eq.~\ref{eq:8} and monitor it through the training of the model on the PACS dataset. The result of this experiment is reported in Figure \ref{fig:sharpness}. Interestingly, CDGA and CDGA$^*$ lead to less sharp local minima, which explains their superior generalization ability.

\section{Conclusions}
\label{sec:Conclusions}

In this paper, to further reduce the estimation error of ERM under domain shift, we proposed CDGA which aims to reduce the shift between all possible domain pairs. CDGA leverages the LDM and generates synthetic images such that the non-iidness in data is reduced. Our results show that CDGA outperforms SOTA DG algorithms under the standard DomainBed benchmark. Our extensive ablation results, including data scaling law, domain shift quantification, adversarial robustness, and loss landscape analysis, shed light on possible reasons that contribute to the success of CDGA in improving ERM in DG setup. Considering the result in Figure \ref{fig:sharpness} for future work it would be interesting to study the theoretical connection of CDGA with the loss landscape structure of the model.

\bibliography{example_paper}
\bibliographystyle{icml2024}

\newpage
\appendix
\onecolumn
\section{Visualization of Different Generative Augmentation Algorithm Results}

\begin{figure*}[ht]
\centering
\includegraphics[width=0.9\textwidth]{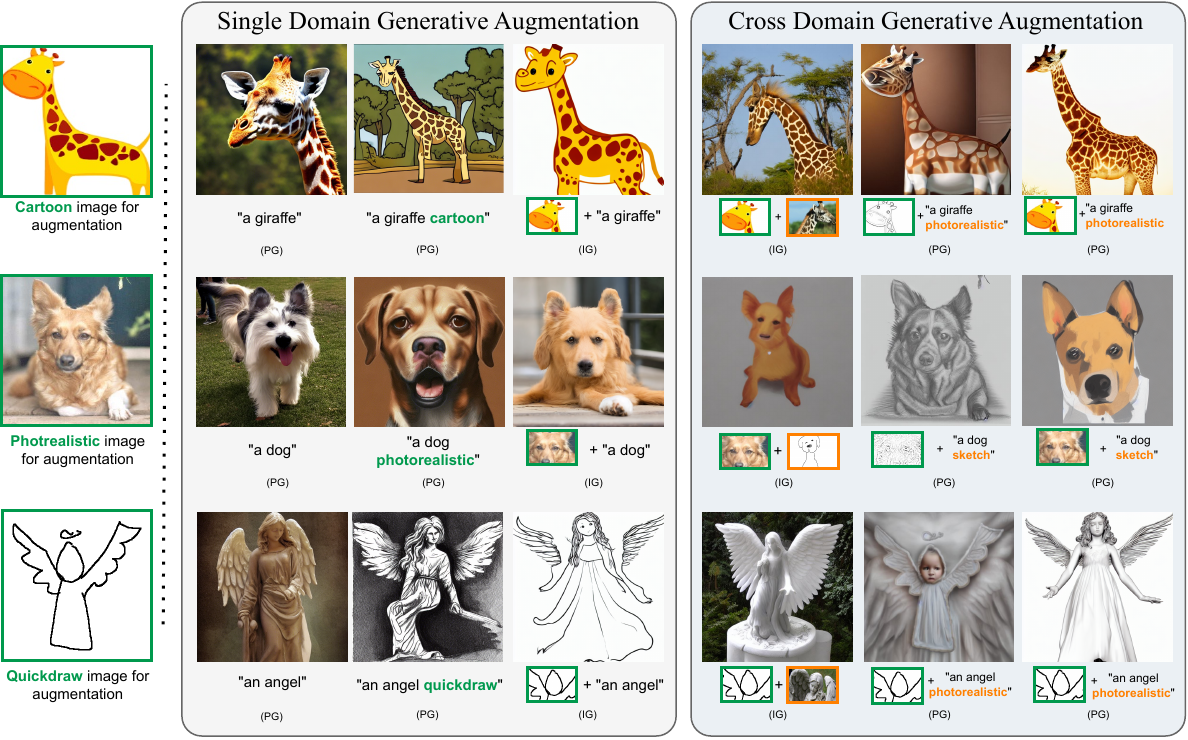} 
\caption{We compare different CDGA and SDGA techniques.  The left column shows the original images intended for generative augmentation. The middle and right columns show different variations of SDGA and CDGA respectively with their prompts. }
\label{fig:aug_samples}
\end{figure*}

In Figure \ref{fig2}, we provided an example that was augmented by different variations of SDGA and CDGA. Figure \ref{fig:aug_samples} is an extension of Figure \ref{fig2} where we present two additional examples and their augmented versions by different variations of SDGA and CDGA.

\section{Test Domain Near-duplication Analysis Full Results}
\begin{figure*}[ht]
\centering
\includegraphics[width=\textwidth]{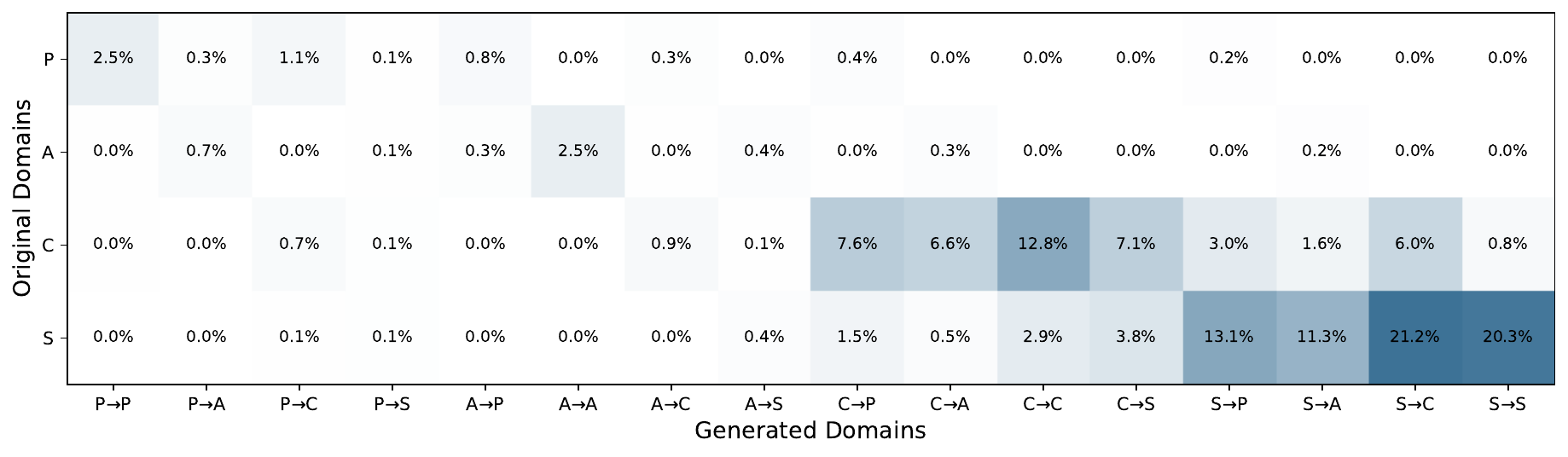} 
\caption{Heat map of the number of near-duplicates of each target domain that are in each original and generated dataset. This table shows that using test-domain description results in more near-duplicate images.}
\label{fig:dup_heatmap_full}
\end{figure*}

\begin{figure*}[ht]
\centering
\includegraphics[width=\textwidth]{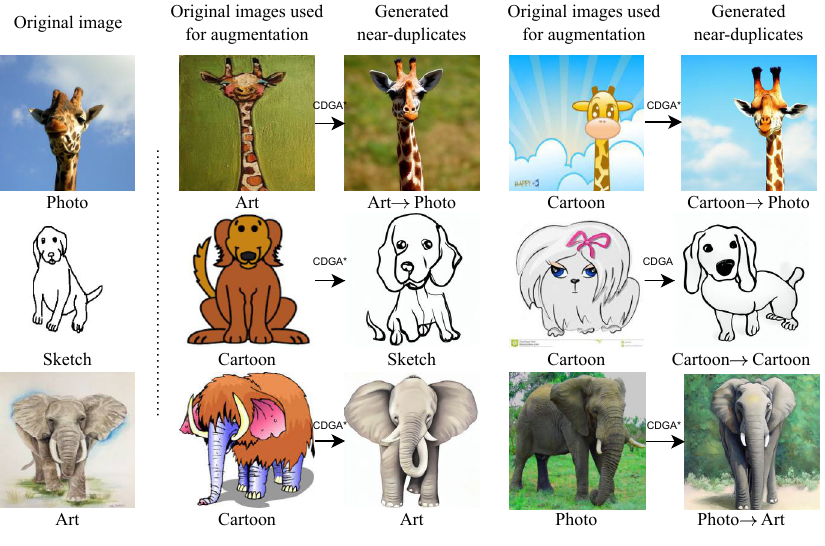} 
\caption{Illustration of near-duplicates of three images from the test domain (left-most column) that are generated using cross-domain generative augmentation (denoted by the arrow) from the original images and are in the training domain.}
\label{fig:near_dup_illus_full}
\end{figure*}

To quantify how much the generated images are similar to the original images for each domain, in section \ref{sec:Results}, Figures \ref{fig:dup_heatmap} and \ref{fig:near_dup_illus} we presented the summarised results for near-duplicate image detection. More precisely, near-duplicate image detection was applied to images generated using CDGA to quantify how much the generated images are similar to the original images for each domain. Here, we present the extended version of these results in Figures \ref{fig:dup_heatmap_full} and
\ref{fig:near_dup_illus_full} respectively. In Figure \ref{fig:dup_heatmap_full}, for each original domain, we report the percentage of near-duplicates over the size of the original domain. Clearly, generating synthetic images that exist in the manifold between training domains enables us to have examples near-duplicate to the target domain. Figure \ref{fig:near_dup_illus_full}  shows multiple examples where the synthetically generated images are near-duplicates to real data. These examples show how CDGA can reduce the domain shift between training domains and the target domain.

\section{Transferability Full Results}

In section \ref{sec:Results}, we monitored classifier heads' Hessian distances as a measure of domain shift and transferability. We calculated classifier heads' Hessian distances between all possible domain pairs through the training steps for ERM, CDGA, and CDGA$^*$ where we set domains P, A, and C as train (source) and domain S as the target. Figure \ref{fig:Hessian_diff}, shows the difference between the classifier head's Hessians given data from domains A and S during the steps. For the sake of completeness, here we extend this result and present classifier heads' Hessian distances between all domain pairs namely (A,S), (P,S), and (C,S). Clearly, in all cases, The distance between classifier heads for CDGA and  CDGA$^*$ is smaller compared with the case where only real data is used (ERM).

\begin{figure*}[ht]
  \centering
  \medskip
  \begin{subfigure}[t]{0.32\linewidth}
    \centering\includegraphics[width=\columnwidth]{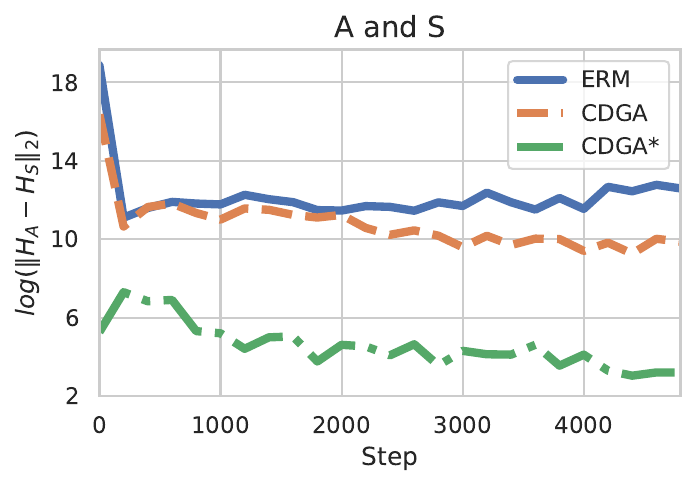}
    \caption{}
  \end{subfigure}
  \hfill
  \begin{subfigure}[t]{0.32\linewidth}
    \centering\includegraphics[width=\columnwidth]{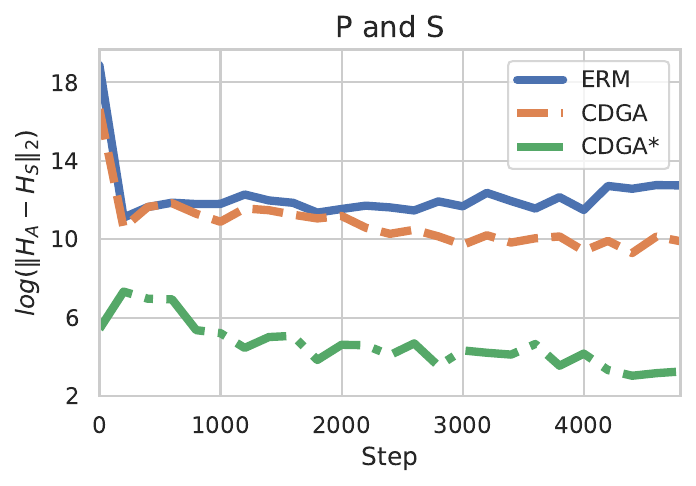}
    \caption{}
    \end{subfigure}
    \hfill
  \begin{subfigure}[t]{0.32\linewidth}
    \centering\includegraphics[width=\columnwidth]{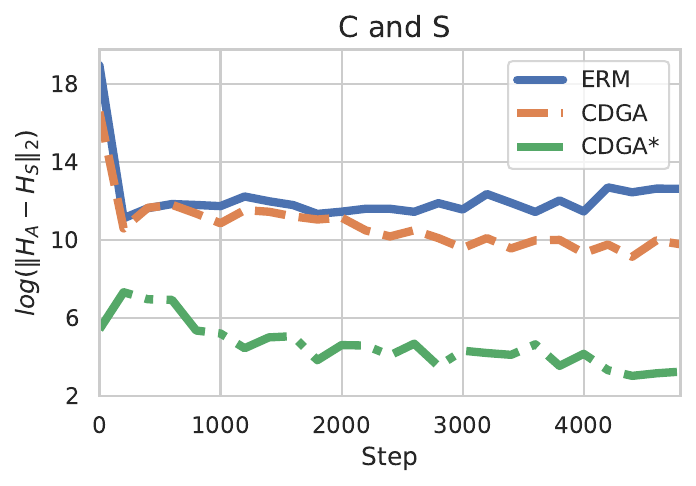}
    \caption{}
    \end{subfigure}
\caption{Classifier head Hessian difference during training for PACS and Cross Domain generative Augmented PACS datasets i.e., CDGA-PACS and CDGA$^*$-PACS where Domains A P, and C are training domains and domain S is the target domain.}
  \label{fig:Hessian_diff_full}%
\end{figure*}

\section{$t$-SNE plots}

\begin{figure*}[ht]
  \centering
  \begin{subfigure}[t]{0.32\linewidth}
    \centering\includegraphics[width=\columnwidth]{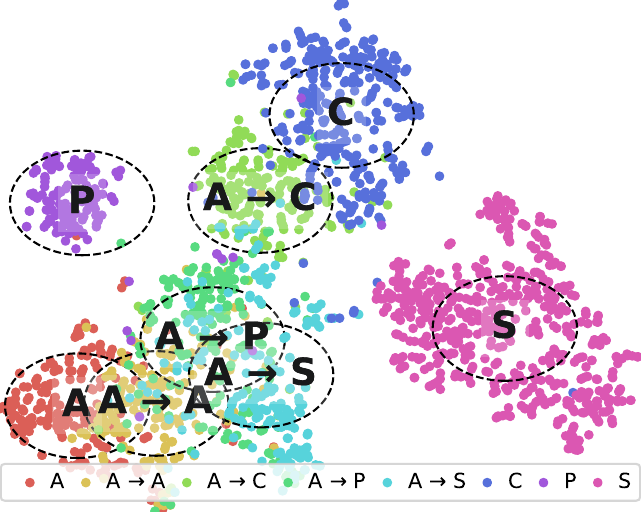}
    \caption{elephant}
  \end{subfigure}
  \hfill
  \begin{subfigure}[t]{0.32\linewidth}
    \centering\includegraphics[width=\columnwidth]{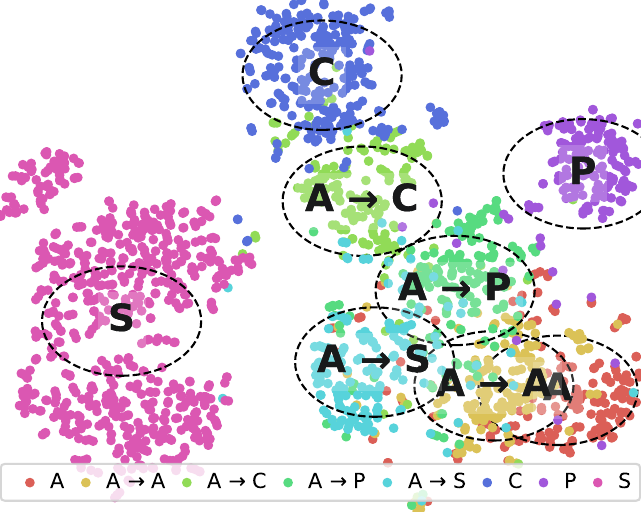}
    \caption{horse}
    \end{subfigure}
    \hfill
  \begin{subfigure}[t]{0.32\linewidth}
    \centering\includegraphics[width=\columnwidth]{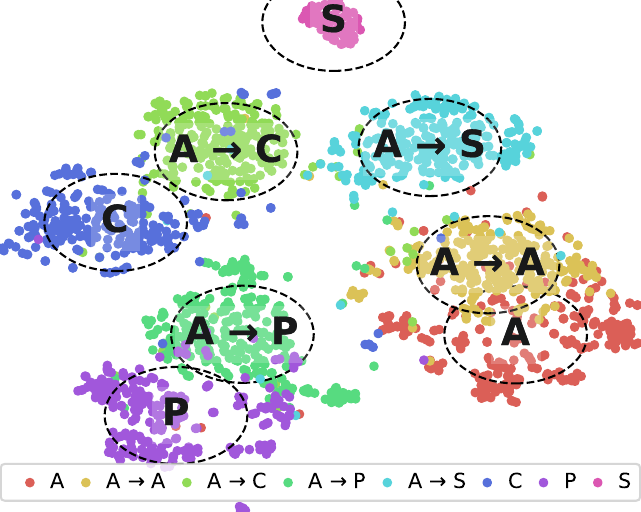}
    \caption{house}
    \end{subfigure}
    \\
    \vspace{20pt}
      \begin{subfigure}[t]{0.32\linewidth}
    \centering\includegraphics[width=\columnwidth]{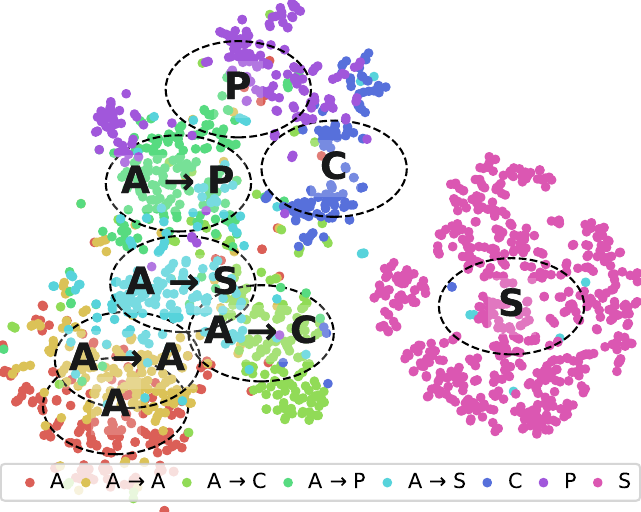}
    \caption{guitar}
  \end{subfigure}
  \hfill
  \begin{subfigure}[t]{0.32\linewidth}
    \centering\includegraphics[width=\columnwidth]{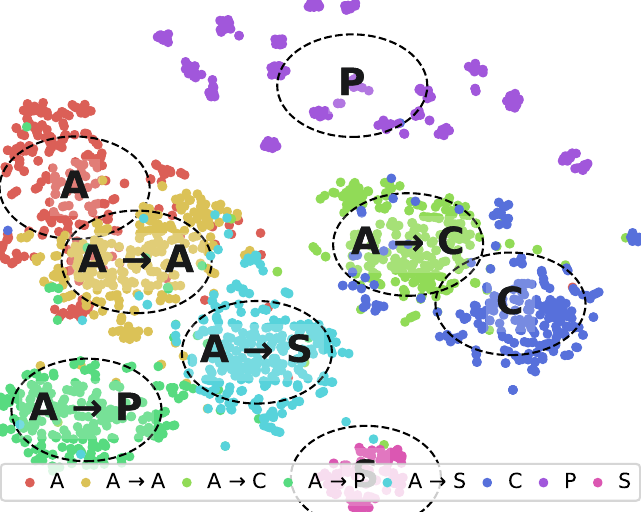}
    \caption{person}
    \end{subfigure}
    \hfill
  \begin{subfigure}[t]{0.32\linewidth}
    \centering\includegraphics[width=\columnwidth]{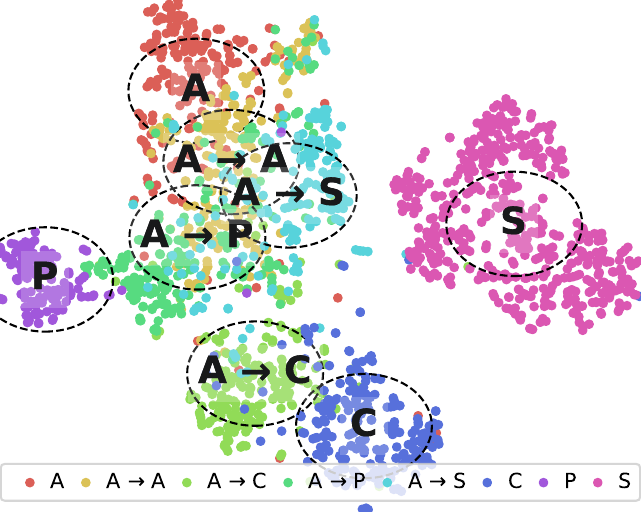}
    \caption{giraffe}
    \end{subfigure}
\caption{The $t$-SNE plot of features extracted from the original PACS dataset and generated images using CDGA by the LDM from A domain for all classes. This figure shows that CDGA can fill the gap between domains.}
  \label{fig:tsne_full}%
\end{figure*}

In Figure \ref{fig:tsne}, we presented a 2D projection of the original PACS dataset from all domains along with CDGA-based data obtained from Domain A only for the ``Dog" class. This figure showed how the cross-domain synthetic images interpolate different domains as we desired. Here in Figure \ref{fig:tsne_full}, we present the results of this experiment for all other classes in the PACS dataset. As can be seen, for most classes the synthetic examples consistently reduce the domain shift which results in better OOD performance of ERM.

\section{DomainBed benchmark full results}

To save space in the main paper, for the DomainBed results in Tables \ref{table:Table1}- \ref{table:Table4} we only reported the five top-performing methods for each model selection technique. Here in Tables \ref{table:Table1_all_methods}, \ref{table:Table2_all_methods}, \ref{table:Table3_all_methods}, and \ref{table:Table4_all_methods} we present the results for all algorithms that have been tested on the DomainBed benchmark \citep{rame2022fishr,gulrajani2020search}. Given that all the results presented for the DomainBed so far are averaged performances for the leave-one-domain-out experiments. The detailed per-domain results for PACS, OfficeHome, DomainNet, and VLCS are presented in Tables \ref{table:Table_PACS_training_full}, \ref{table:Table_PACS_validation_full}, \ref{table:Table_PACS_oracle_full}, \ref{table:Table_Officehome_training_full}, \ref{table:Table_Officehome_validation_full}, \ref{table:Table_Officehome_oracle_full}, \ref{table:Table_DomainNet_training_full}, \ref{table:Table_DomainNet_validation_full}, \ref{table:Table_DomainNet_oracle_full}, 
\ref{table:Table_VLCS_training_full}, \ref{table:Table_VLCS_validation_full}, and \ref{table:Table_VLCS_oracle_full}.

\begin{table*}[ht]%
    \caption{DomainBed benchmark for \textbf{training-domain validation set} model selection method. We format \textbf{first}, \underline{second} and \textcolor{gray}{worse than ERM} results.}
    \centering
    {
        \begin{tabular}{l|cccc}
            \toprule
\multirow{3}{*}{\textbf{Algorithm}} & \multicolumn{4}{c}{}                                                                                                                                                                                                                  \\
                                       & \textbf{PACS}                                 & \textbf{OfficeHome}  & \textbf{DomainNet}                                                      & \textbf{Avg}               \\
            \midrule
            ERM                                                           & 85.5 \scriptsize{$\pm$ 0.2}                   & 66.5 \scriptsize{$\pm$ 0.3}
            & 40.9 \scriptsize{$\pm$ 1.8}                                   & 64.3                               \\
            IRM                                           & \textcolor{gray}{83.5} \scriptsize{$\pm$ 0.8} & \textcolor{gray}{64.3} \scriptsize{$\pm$ 2.2}                    & \textcolor{gray}{33.9} \scriptsize{$\pm$ 2.8} & \textcolor{gray}{60.6 }  \\
            GroupDRO                                  & \textcolor{gray}{84.4} \scriptsize{$\pm$ 0.8} & \textcolor{gray}{66.0} \scriptsize{$\pm$ 0.7} & \textcolor{gray}{33.3} \scriptsize{$\pm$ 0.2} &\textcolor{gray}{61.2}\\
            Mixup                                   & \textcolor{gray}{84.6} \scriptsize{$\pm$ 0.6} & 68.1 \scriptsize{$\pm$ 0.3}                       & \textcolor{gray}{39.2} \scriptsize{$\pm$ 0.1} & \textcolor{gray}{64.0}              \\
            MLDG                                 & \textcolor{gray}{84.9} \scriptsize{$\pm$ 1.0} & 66.8 \scriptsize{$\pm$ 0.6}                                   & 41.2 \scriptsize{$\pm$ 0.1}                   & 64.3          \\
            CORAL                                                  & 86.2 \scriptsize{$\pm$ 0.3}       & \underline{68.7} \scriptsize{$\pm$ 0.3}                      & 41.5 \scriptsize{$\pm$ 0.1}                   & 65.5           \\
            MMD                                                                & \textcolor{gray}{84.6} \scriptsize{$\pm$ 0.5} & \textcolor{gray}{66.3} \scriptsize{$\pm$ 0.1}  & \textcolor{gray}{23.4} \scriptsize{$\pm$ 9.5} & \textcolor{gray}{58.1}  \\
            DANN                                                                & \textcolor{gray}{83.6} \scriptsize{$\pm$ 0.4} & \textcolor{gray}{65.9} \scriptsize{$\pm$ 0.6}             & \textcolor{gray}{38.3} \scriptsize{$\pm$ 0.1} & \textcolor{gray}{62.6}  \\
            CDANN                                                              & \textcolor{gray}{82.6} \scriptsize{$\pm$ 0.9} & \textcolor{gray}{65.8} \scriptsize{$\pm$ 1.3}  & \textcolor{gray}{38.3} \scriptsize{$\pm$ 0.3} & \textcolor{gray}{62.2}  \\
            MTL                                & \textcolor{gray}{84.6} \scriptsize{$\pm$ 0.5} & \textcolor{gray}{66.4} \scriptsize{$\pm$ 0.5} & \textcolor{gray}{40.6} \scriptsize{$\pm$ 0.1} & \textcolor{gray}{63.9}  \\
            SagNet                                                            & 86.3 \scriptsize{$\pm$ 0.2}          & 68.1 \scriptsize{$\pm$ 0.1}                      & \textcolor{gray}{40.3} \scriptsize{$\pm$ 0.1} & 64.9        \\
            ARM                                                          & \textcolor{gray}{85.1} \scriptsize{$\pm$ 0.4} & \textcolor{gray}{64.8} \scriptsize{$\pm$ 0.3}  & \textcolor{gray}{35.5} \scriptsize{$\pm$ 0.2} & \textcolor{gray}{61.8}  \\
            V-REx                                                        & \textcolor{gray}{84.9} \scriptsize{$\pm$ 0.6} & \textcolor{gray}{66.4} \scriptsize{$\pm$ 0.6}                & \textcolor{gray}{33.6} \scriptsize{$\pm$ 2.9} & \textcolor{gray}{61.6}  \\
            RSC                                                  & \textcolor{gray}{85.2} \scriptsize{$\pm$ 0.9} & \textcolor{gray}{65.5} \scriptsize{$\pm$ 0.9}                  & \textcolor{gray}{38.9} \scriptsize{$\pm$ 0.5} & \textcolor{gray}{63.2}  \\
            AND-mask                                    & \textcolor{gray}{84.4} \scriptsize{$\pm$ 0.9} & \textcolor{gray}{65.6} \scriptsize{$\pm$ 0.4}  & \textcolor{gray}{37.2} \scriptsize{$\pm$ 0.6} & \textcolor{gray}{62.4}  \\
            SAND-mask                  & \textcolor{gray}{84.6} \scriptsize{$\pm$ 0.9} & \textcolor{gray}{65.8} \scriptsize{$\pm$ 0.4}  & \textcolor{gray}{32.1} \scriptsize{$\pm$ 0.6} & \textcolor{gray}{60.8}  \\
            Fish                                                                & 85.5 \scriptsize{$\pm$ 0.3}                   & 68.6 \scriptsize{$\pm$ 0.4}        & 42.7 \scriptsize{$\pm$ 0.2}          & 65.6                \\
        
            Fishr                                              & 85.5 \scriptsize{$\pm$ 0.4}           & 67.8 \scriptsize{$\pm$ 0.1}                                  & 41.7 \scriptsize{$\pm$ 0.0}       & 65.0                             
                          \\ 
             HGP                                         & 84.7 \scriptsize{$\pm$ 0.0}           & 68.2 \scriptsize{$\pm$ 0.0}           &41.1 \scriptsize{$\pm$ 0.0}   & 64.7                \\
 Hutchinson     & 83.9 \scriptsize{$\pm$ 0.0}           & 68.2 \scriptsize{$\pm$ 0.0}   &41.6 \scriptsize{$\pm$ 0.0}          &   64.6  \\

                \midrule

    CDGA-PG                 & \underline{88.5}\scriptsize{ $\pm$ 0.5}    &  68.2 \scriptsize{$\pm$  0.6    }                                  & \underline{43.1} \scriptsize{$\pm$0.0}       & \underline{66.6}   \\

      CDGA-PG$^*$                   & \textbf{89.5}\scriptsize{$\pm$ 0.3}            & \textbf{70.8} \scriptsize{$\pm$ 0.6 }                                 & \textbf{44.8} \scriptsize{$\pm$0.0}       & \textbf{68.4}  \\

     \bottomrule
        \end{tabular}
    }
    \label{table:Table1_all_methods}%
\end{table*}

          

\begin{table*}[ht]%
    \caption{DomainBed benchmark for \textbf{leave-one-domain-out cross-validation} model selection. We format \textbf{first}, \underline{second} and \textcolor{gray}{worse than ERM} results.}%
    \centering
    {
\begin{tabular}{l|ccccc}
            \toprule
            \multirow{3}{*}{\textbf{Algorithm}} & \multicolumn{4}{c}{}                                                                                                                                                                                                                                                                                                                                                               \\
  & \textbf{PACS}                                 & \textbf{OfficeHome}   & \textbf{DomainNet}                                                     & \textbf{Avg}               \\
\midrule
            ERM                                                  & 83.0  \scriptsize{$\pm$ 0.7}            & 65.7 
  \scriptsize{$\pm$ 0.5}                 & 40.6 \scriptsize{$\pm$ 0.2}         & 63.1                    \\
            IRM                                                & \textcolor{gray}{81.5} \scriptsize{$\pm$ 0.8} & \textcolor{gray}{64.3} \scriptsize{$\pm$ 1.5}  &\textcolor{gray}{33.5} \scriptsize{$\pm$ 0.3}  & \textcolor{gray}{59.8}  \\
            GroupDRO                            & 83.5  \scriptsize{$\pm$ 0.2   } & \textcolor{gray}{65.2} \scriptsize{$\pm$ 0.2   }    & \textcolor{gray}{33.0} \scriptsize{$\pm$ 0.3}      & \textcolor{gray}{60.6}          \\
            Mixup                      & {83.2} \scriptsize{$\pm$ 0.4}     & 67.0 \scriptsize{$\pm$ 0.2}        & \textcolor{gray}{38.5} \scriptsize{$\pm$ 0.3} & \textcolor{gray}{62.9}                                   \\
            MLDG                                     & \textcolor{gray}{82.9} \scriptsize{$\pm$ 1.7} & 66.1 \scriptsize{$\pm$ 0.5}      & 41.0 \scriptsize{$\pm$ 0.2}                                    & 63.3                   \\
\bottomrule
        \end{tabular}
    }
    \label{table:Table2_all_methods}%
\end{table*}

\begin{table*}[ht]%
    \caption{DomainBed benchmark for \textbf{test-domain validation set (oracle)} model selection method. We format \textbf{first}, \underline{second} and \textcolor{gray}{worse than ERM} results.}%
    \centering
    {
        \begin{tabular}{l|cccc}
            \toprule
            \multirow{3}{*}{\textbf{Algorithm}} & \multicolumn{4}{c}{}                                                                                                                                                                                                                                                                                                                                                               \\                                  & \textbf{PACS}     &\textbf{OfficeHome}  & \textbf{DomainNet}                                                      & \textbf{Avg}               \\
            \midrule
            ERM                                                                    & 86.7 \scriptsize{$\pm$ 0.3}                   & 66.4 \scriptsize{$\pm$ 0.5}                               & 41.3 \scriptsize{$\pm$ 0.1}                   & 64.8             \\
            IRM                                  & \textcolor{gray}{84.5} \scriptsize{$\pm$ 1.1} & \textcolor{gray}{63.0} \scriptsize{$\pm$ 2.7}  & \textcolor{gray}{28.0} \scriptsize{$\pm$ 5.1} & \textcolor{gray}{58.5}   \\
            
            GroupDRO                                                                                & 87.1 \scriptsize{$\pm$ 0.1}                   & \textcolor{gray}{66.2} \scriptsize{$\pm$ 0.6}  & \textcolor{gray}{33.4} \scriptsize{$\pm$ 0.3} & \textcolor{gray}{62.2}\\
            Mixup                                                                 & 86.8 \scriptsize{$\pm$ 0.3}                   & 68.0 \scriptsize{$\pm$ 0.2}                          & \textcolor{gray}{39.6} \scriptsize{$\pm$ 0.1}     & 64.8          \\
            MLDG                                                              & 86.8 \scriptsize{$\pm$ 0.4}                   & 66.6 \scriptsize{$\pm$ 0.3}                     & 41.6 \scriptsize{$\pm$ 0.1}                   & 65.0             \\
\bottomrule
        \end{tabular}
    }
    \label{table:Table3_all_methods}%
\end{table*}

\begin{table*}[ht]
\caption{DomainBed benchmark on \textbf{VLCS} dataset across different model selection methods. We format \textbf{first}, \underline{second} and \textcolor{gray}{worse than ERM} results.}

  \centering
  \begin{tabular}{lccc}
    \toprule
     Method & Training domain & Leave-one-domain-out  &Oracle    \\
    \toprule
         ERM &  77.5 ± \scriptsize{0.4}  &  77.2 ± \scriptsize{0.4}  &   77.6 ± \scriptsize{0.3}  \\ 
 IRM                                          & 78.5 \scriptsize{$\pm$ 0.5}    &  \textcolor{gray}{76.3}   \scriptsize{$\pm$ 0.6}   & \textcolor{gray}{76.9} \scriptsize{$\pm$ 0.6}  \\
 
 GroupDRO                            & \textcolor{gray}{76.7} \scriptsize{$\pm$ 0.6}    & 77.9 \scriptsize{$\pm$ 0.5 } &     \textcolor{gray}{77.4} \scriptsize{$\pm$ 0.5}  \\

  Mixup                                & \textcolor{gray}{77.4} \scriptsize{$\pm$ 0.6}    & 77.7 \scriptsize{$\pm$ 0.6  }       &78.1 \scriptsize{$\pm$ 0.3} \\

   MLDG                                 & \textcolor{gray}{77.2} \scriptsize{$\pm$ 0.4}    &     77.2 \scriptsize{$\pm$ 0.9}   & \textcolor{gray}{77.5} \scriptsize{$\pm$ 0.1} \\         

CORAL  & \underline{78.8}  ± \scriptsize{0.6}  &  \underline{78.7} ± \scriptsize{0.4}  &  77.7 ± \scriptsize{0.2}  \\

MMD                                               &  77.5 \scriptsize{$\pm$ 0.9}           & 77.3 \scriptsize{$\pm$ 0.5}      &77.9 \scriptsize{$\pm$ 0.1}    \\
\bottomrule
  \end{tabular}

 \label{table:Table4_all_methods}
\end{table*}

\begin{table*}[ht]%
    \caption{DomainBed benchmark, \textbf{PACS full results for training-domain validation set} model selection method.}%
    \centering
    {
        \begin{tabular}{l|ccccc}
            \toprule
            \multirow{3}{*}{\textbf{Algorithm}} & \multicolumn{4}{c}{}                                                                                                                                                                                                                                                                                                                                                               \\           & \textbf{A}  & \textbf{C}   & \textbf{P}   & \textbf{S}      & \textbf{Avg}      \\
            \midrule
            ERM                & 84.7 $\pm$ 0.4             & 80.8 $\pm$ 0.6    & 97.2 $\pm$ 0.3             & 79.3 $\pm$ 1.0 & 85.5                           \\
            IRM                & 84.8 $\pm$ 1.3             & 76.4 $\pm$ 1.1             & 96.7 $\pm$ 0.6             & 76.1 $\pm$ 1.0             & 83.5                          \\
            GroupDRO           & 83.5 $\pm$ 0.9             & 79.1 $\pm$ 0.6             & 96.7 $\pm$ 0.3             & 78.3 $\pm$ 2.0             & 84.4                            \\
            Mixup              & 86.1 $\pm$ 0.5             & 78.9 $\pm$ 0.8             & 97.6 $\pm$ 0.1             & 75.8 $\pm$ 1.8             & 84.6                            \\
            MLDG               & 85.5 $\pm$ 1.4             & 80.1 $\pm$ 1.7             & 97.4 $\pm$ 0.3             & 76.6 $\pm$ 1.1             & 84.9                           \\
            CORAL              & 88.3 $\pm$ 0.2             & 80.0 $\pm$ 0.5             & \underline{97.5} $\pm$ 0.3 & 78.8 $\pm$ 1.3             & 86.2              \\
\end{tabular}
    }
    \label{table:Table_PACS_training_full}%
\end{table*}

\begin{table*}[ht]%
    \caption{DomainBed benchmark, \textbf{PACS full results for leave-one-domain-out cross-validation}  model selection method.}%
    \centering
    {
        \begin{tabular}{l|ccccc}
            \toprule
            \multirow{3}{*}{\textbf{Algorithm}} & \multicolumn{4}{c}{}                                                                                                                                                                                                                                                                                                                                                               \\           & \textbf{A}  & \textbf{C}   & \textbf{P}   & \textbf{S}      & \textbf{Avg}      \\
            \midrule
   ERM                  & 83.2 $\pm$ 1.3       & 76.8 $\pm$ 1.7       & \textbf{97.2} $\pm$ 0.3       & 74.8 $\pm$ 1.3       & 83.0                  \\
    IRM                  & 81.7 $\pm$ 2.4       & 77.0 $\pm$ 1.3       & 96.3 $\pm$ 0.2       & 71.1 $\pm$ 2.2       & 81.5                \\
    GroupDRO             & 84.4 $\pm$ 0.7       & 77.3 $\pm$ 0.8       & 96.8 $\pm$ 0.8       & 75.6 $\pm$ 1.4       & 83.5                 \\
    Mixup                & 85.2 $\pm$ 1.9       & 77.0 $\pm$ 1.7       & 96.8 $\pm$ 0.8       & 73.9 $\pm$ 1.6       & 83.2                 \\
    MLDG                 & 81.4 $\pm$ 3.6       & 77.9 $\pm$ 2.3       & 96.2 $\pm$ 0.3       & 76.1 $\pm$ 2.1       & 82.9                 \\
    CORAL                & 80.5 $\pm$ 2.8       & 74.5 $\pm$ 0.4       & 96.8 $\pm$ 0.3       & 78.6 $\pm$ 1.4       & 82.6               \\
\end{tabular}
    }
    \label{table:Table_PACS_validation_full}%
\end{table*}

\begin{table*}[ht]%
    \caption{DomainBed benchmark, \textbf{PACS full results for test-domain validation set (oracle)} model selection method.}%
    \centering
    {
        \begin{tabular}{l|ccccc}
            \toprule
            \multirow{3}{*}{\textbf{Algorithm}} & \multicolumn{4}{c}{}                                                                                                                                                                                                                                                                                                                                                               \\           & \textbf{A}  & \textbf{C}   & \textbf{P}   & \textbf{S}      & \textbf{Avg}      \\
            \midrule
   ERM                & 86.5 $\pm$ 1.0             & 81.3 $\pm$ 0.6             & 96.2 $\pm$ 0.3             & 82.7 $\pm$ 1.1 & 86.7                          \\
    IRM                & 84.2 $\pm$ 0.9             & 79.7 $\pm$ 1.5             & 95.9 $\pm$ 0.4             & 78.3 $\pm$ 2.1          & 84.5                         \\
    GroupDRO           & 87.5 $\pm$ 0.5             & 82.9 $\pm$ 0.6    & 97.1 $\pm$ 0.3             & 81.1 $\pm$ 1.2          & 87.1                         \\
    Mixup              & 87.5 $\pm$ 0.4             & 81.6 $\pm$ 0.7             & \underline{97.4} $\pm$ 0.2 & 80.8 $\pm$ 0.9          & 86.8                          \\
    MLDG               & 87.0 $\pm$ 1.2             & 82.5 $\pm$ 0.9             & 96.7 $\pm$ 0.3             & 81.2 $\pm$ 0.6          & 86.8                          \\
    CORAL              & 86.6 $\pm$ 0.8             & 81.8 $\pm$ 0.9             & 97.1 $\pm$ 0.5             & 82.7 $\pm$ 0.6 & 87.1                          \\
\end{tabular}
    }
    \label{table:Table_PACS_oracle_full}%
\end{table*}

\begin{table*}[ht]%
    \caption{DomainBed benchmark, \textbf{OfficeHome full results for training-domain validation set} model selection method.}%
    \centering
    {
        \begin{tabular}{l|ccccc}
            \toprule
            \multirow{3}{*}{\textbf{Algorithm}} & \multicolumn{4}{c}{}                                                                                                                                                                                                                                                                                                                                                               \\           & \textbf{A}  & \textbf{C}   & \textbf{P}   & \textbf{R}      & \textbf{Avg}      \\
            \midrule
                       ERM                & 61.3 $\pm$ 0.7             & 52.4 $\pm$ 0.3          & 75.8 $\pm$ 0.1             & 76.6 $\pm$ 0.3             & 66.5                             \\
            IRM                & 58.9 $\pm$ 2.3             & 52.2 $\pm$ 1.6          & 72.1 $\pm$ 2.9             & 74.0 $\pm$ 2.5             & 64.3                           \\
            GroupDRO           & 60.4 $\pm$ 0.7             & 52.7 $\pm$ 1.0          & 75.0 $\pm$ 0.7             & 76.0 $\pm$ 0.7             & 66.0                            \\
            Mixup              & 62.4 $\pm$ 0.8             & \underline{54.8} $\pm$ 0.6 & \textbf{76.9} $\pm$ 0.3    & 78.3 $\pm$ 0.2 & 68.1                             \\
            MLDG               & 61.5 $\pm$ 0.9             & 53.2 $\pm$ 0.6          & 75.0 $\pm$ 1.2             & 77.5 $\pm$ 0.4             & 66.8                            \\
            CORAL              & \textbf{65.3} $\pm$ 0.4    & 54.4 $\pm$ 0.5          & \underline{76.5} $\pm$ 0.1 & 78.4 $\pm$ 0.5    & \underline{{68.7}}                   \\
\end{tabular}
        }
    \label{table:Table_Officehome_training_full}%
\end{table*}

\begin{table*}[ht]%
    \caption{DomainBed benchmark, \textbf{OfficeHome full results for leave-one-domain-out cross-validation} model selection method.}%
    \centering
    {
        \begin{tabular}{l|ccccc}
            \toprule
            \multirow{3}{*}{\textbf{Algorithm}} & \multicolumn{4}{c}{}                                  \\           & \textbf{A}  & \textbf{C}   & \textbf{P}   & \textbf{R}      & \textbf{Avg}      \\
            \midrule
 ERM                  & 61.1 $\pm$ 0.9       & 50.7 $\pm$ 0.6       & 74.6 $\pm$ 0.3       & 76.4 $\pm$ 0.6       & 65.7                 \\
IRM                  & 58.2 $\pm$ 1.2       & 51.6 $\pm$ 1.2       & 73.3 $\pm$ 2.2       & 74.1 $\pm$ 1.7       & 64.3                 \\
GroupDRO             & 59.9 $\pm$ 0.4       & 51.0 $\pm$ 0.4       & 73.7 $\pm$ 0.3       & 76.0 $\pm$ 0.2       & 65.2                 \\
Mixup                & 61.4 $\pm$ 0.5       & 53.0 $\pm$ 0.3       & 75.8 $\pm$ 0.2       & 77.7 $\pm$ 0.3       & 67.0                \\
MLDG                 & 60.5 $\pm$ 1.4       & 51.9 $\pm$ 0.2       & 74.4 $\pm$ 0.6       & 77.6 $\pm$ 0.4       & 66.1                \\
CORAL                & 64.5 $\pm$ 0.8       & 54.8 $\pm$ 0.2       & 76.6 $\pm$ 0.3       & 78.1 $\pm$ 0.2       & 68.5                  \\
\end{tabular}
    }
\label{table:Table_Officehome_validation_full}%
\end{table*}

\begin{table*}[ht]%
    \caption{DomainBed benchmark, \textbf{OfficeHome full results for test-domain validation set (oracle)} model selection method.}%
    \centering
    {
        \begin{tabular}{l|ccccc}
            \toprule
            \multirow{3}{*}{\textbf{Algorithm}} & \multicolumn{4}{c}{}                                                                                                                                                                                                                                                                                                                                                               \\           & \textbf{A}  & \textbf{C}   & \textbf{P}   & \textbf{R}      & \textbf{Avg}      \\
            \midrule
        ERM                & 61.7 $\pm$ 0.7             & 53.4 $\pm$ 0.3             & 74.1 $\pm$ 0.4             & 76.2 $\pm$ 0.6             & 66.4                          \\
        IRM                & 56.4 $\pm$ 3.2             & 51.2 $\pm$ 2.3             & 71.7 $\pm$ 2.7             & 72.7 $\pm$ 2.7             & 63.0                           \\
        GroupDRO           & 60.5 $\pm$ 1.6             & 53.1 $\pm$ 0.3             & 75.5 $\pm$ 0.3             & 75.9 $\pm$ 0.7             & 66.2                            \\
        Mixup              & \underline{63.5} $\pm$ 0.2 & 54.6 $\pm$ 0.4    & 76.0 $\pm$ 0.3             & 78.0 $\pm$ 0.7             & 68.0                           \\
        MLDG               & 60.5 $\pm$ 0.7             & 54.2 $\pm$ 0.5 & 75.0 $\pm$ 0.2             & 76.7 $\pm$ 0.5             & 66.6                            \\
        CORAL              & \textbf{64.8} $\pm$ 0.8    & 54.1 $\pm$ 0.9             & 76.5 $\pm$ 0.4    & 78.2 $\pm$ 0.4 & 68.4                    \\
\end{tabular}
    }
    \label{table:Table_Officehome_oracle_full}%
\end{table*}

\begin{table*}[ht]%
    \caption{DomainBed benchmark, \textbf{DomainNet full results for training-domain validation set} model selection method.}%
    \centering
    {
        \begin{tabular}{l|cccccccc}
            \toprule
            \multirow{3}{*}{\textbf{Algorithm}} & \multicolumn{7}{c}{}                                                                                                                                                                                                                                                                                                                                                               \\           & \textbf{clip}  & \textbf{info}   & \textbf{paint}   & \textbf{quick}   & \textbf{real} & \textbf{sketch}   & \textbf{Avg}     \\
            \midrule
             ERM                & 58.1 $\pm$ 0.3             & 18.8 $\pm$ 0.3             & 46.7 $\pm$ 0.3 & 12.2 $\pm$ 0.4          & 59.6 $\pm$ 0.1             & 49.8 $\pm$ 0.4             & 40.9                             \\
            IRM                & 48.5 $\pm$ 2.8             & 15.0 $\pm$ 1.5             & 38.3 $\pm$ 4.3             & 10.9 $\pm$ 0.5          & 48.2 $\pm$ 5.2             & 42.3 $\pm$ 3.1             & 33.9                            \\
            GroupDRO           & 47.2 $\pm$ 0.5             & 17.5 $\pm$ 0.4             & 33.8 $\pm$ 0.5             & 9.3 $\pm$ 0.3           & 51.6 $\pm$ 0.4             & 40.1 $\pm$ 0.6             & 33.3                            \\
            Mixup              & 55.7 $\pm$ 0.3             & 18.5 $\pm$ 0.5             & 44.3 $\pm$ 0.5             & 12.5 $\pm$ 0.4          & 55.8 $\pm$ 0.3             & 48.2 $\pm$ 0.5             & 39.2                             \\
            MLDG               & 59.1 $\pm$ 0.2 & 19.1 $\pm$ 0.3             & 45.8 $\pm$ 0.7             & \textbf{13.4} $\pm$ 0.3 & 59.6 $\pm$ 0.2             & 50.2 $\pm$ 0.4 & 41.2                             \\
            CORAL              & 59.2 $\pm$ 0.1    & 19.7 $\pm$ 0.2 & 46.6 $\pm$ 0.3             & \textbf{13.4} $\pm$ 0.4 & 59.8 $\pm$ 0.2 & 50.1 $\pm$ 0.6             & 41.5                            \\
\end{tabular}
    }
    \label{table:Table_DomainNet_training_full}%
\end{table*}

\begin{table*}[ht]%
    \caption{DomainBed benchmark, \textbf{DomainNet full results for leave-one-out} model selection method.}%
    \centering
    {
        \begin{tabular}{l|cccccccc}
            \toprule
            \multirow{3}{*}{\textbf{Algorithm}} & \multicolumn{7}{c}{}                                                                                                                                                                                                                                                                                                                                                               \\           & \textbf{clip}  & \textbf{info}   & \textbf{paint}   & \textbf{quick}   & \textbf{real} & \textbf{sketch}   & \textbf{Avg}     \\
            \midrule
        ERM                  & 58.1 $\pm$ 0.3       & 17.8 $\pm$ 0.3       & 47.0 $\pm$ 0.3       & 12.2 $\pm$ 0.4       & 59.2 $\pm$ 0.7       & 49.5 $\pm$ 0.6       & 40.6           \\
        IRM                  & 47.5 $\pm$ 2.7       & 15.0 $\pm$ 1.5       & 37.3 $\pm$ 5.1       & 10.9 $\pm$ 0.5       & 48.0 $\pm$ 5.4       & 42.3 $\pm$ 3.1       & 33.5                 \\
        GroupDRO             & 47.2 $\pm$ 0.5       & 17.0 $\pm$ 0.6       & 33.8 $\pm$ 0.5       & 9.2 $\pm$ 0.4        & 51.6 $\pm$ 0.4       & 39.2 $\pm$ 1.2       & 33.0              \\
        Mixup                & 54.4 $\pm$ 0.6       & 18.0 $\pm$ 0.4       & 44.5 $\pm$ 0.5       & 11.5 $\pm$ 0.2       & 55.8 $\pm$ 1.1       & 46.9 $\pm$ 0.2       & 38.5              \\
        MLDG                 & 58.3 $\pm$ 0.7       & 19.3 $\pm$ 0.2       & 45.8 $\pm$ 0.7       & \underline{13.2} $\pm$ 0.3       & 59.4 $\pm$ 0.2       & 49.8 $\pm$ 0.3       & 41.0                \\
        CORAL                & 59.2 $\pm$ 0.1       & 19.5 $\pm$ 0.3       & 46.2 $\pm$ 0.1       & \textbf{13.4} $\pm$ 0.4       & 59.1 $\pm$ 0.5       & 49.5 $\pm$ 0.8       & 41.1               \\
\end{tabular}
    }
    \label{table:Table_DomainNet_validation_full}%
\end{table*}

\begin{table*}[ht]%
    \caption{DomainBed benchmark, \textbf{DomainNet full results for test-domain validation set} (oracle) model selection method.}%
    \centering
    {
        \begin{tabular}{l|cccccccc}
            \toprule
            \multirow{3}{*}{\textbf{Algorithm}} & \multicolumn{7}{c}{}                                                                                                                                                                                                                                                                                                                                                               \\           & \textbf{clip}  & \textbf{info}   & \textbf{paint}   & \textbf{quick}   & \textbf{real} & \textbf{sketch}   & \textbf{Avg}      \\
            \midrule
ERM                & 58.6 $\pm$ 0.3             & 19.2 $\pm$ 0.2             & 47.0 $\pm$ 0.3             & 13.2 $\pm$ 0.2             & 59.9 $\pm$ 0.3             & 49.8 $\pm$ 0.4             & 41.3                           \\
            IRM                & 40.4 $\pm$ 6.6             & 12.1 $\pm$ 2.7             & 31.4 $\pm$ 5.7             & 9.8 $\pm$ 1.2              & 37.7 $\pm$ 9.0             & 36.7 $\pm$ 5.3             & 28.0                           \\
            GroupDRO           & 47.2 $\pm$ 0.5             & 17.5 $\pm$ 0.4             & 34.2 $\pm$ 0.3             & 9.2 $\pm$ 0.4              & 51.9 $\pm$ 0.5             & 40.1 $\pm$ 0.6             & 33.4                            \\
            Mixup              & 55.6 $\pm$ 0.1             & 18.7 $\pm$ 0.4             & 45.1 $\pm$ 0.5             & 12.8 $\pm$ 0.3             & 57.6 $\pm$ 0.5             & 48.2 $\pm$ 0.4             & 39.6                             \\
            MLDG               & 59.3 $\pm$ 0.1    & 19.6 $\pm$ 0.2             & 46.8 $\pm$ 0.2             & 13.4 $\pm$ 0.2             & 60.1 $\pm$ 0.4 & 50.4 $\pm$ 0.3 & 41.6                             \\
            CORAL              & 59.2 $\pm$ 0.1 & 19.9 $\pm$ 0.2 & 47.4 $\pm$ 0.2 & \textbf{14.0} $\pm$ 0.4    & 59.8 $\pm$ 0.2             & 50.4 $\pm$ 0.4 & 41.8                 \\
\end{tabular}
    }
    \label{table:Table_DomainNet_oracle_full}%
\end{table*}

\begin{table*}[ht]%
    \caption{DomainBed benchmark, \textbf{VLCS full results for training-domain validation set} model selection method.}%
    \centering
    {
        \begin{tabular}{l|ccccc}
            \toprule
            \multirow{3}{*}{\textbf{Algorithm}} & \multicolumn{4}{c}{}                                                                                                                                                                                                                                                                                                                                                               \\           & \textbf{C}  & \textbf{L}   & \textbf{S}   & \textbf{V}      & \textbf{Avg}     \\
            \midrule
            ERM                & 97.7 $\pm$ 0.4             & 64.3 $\pm$ 0.9             & 73.4 $\pm$ 0.5             & 74.6 $\pm$ 1.3          & 77.5                            \\
            IRM                & 98.6 $\pm$ 0.1             & 64.9 $\pm$ 0.9             & 73.4 $\pm$ 0.6             & \underline{77.3} $\pm$ 0.9          & 78.5                 \\
            GroupDRO           & 97.3 $\pm$ 0.3             & 63.4 $\pm$ 0.9             & 69.5 $\pm$ 0.8             & 76.7 $\pm$ 0.7          & 76.7                            \\
            Mixup              & 98.3 $\pm$ 0.6             & 64.8 $\pm$ 1.0             & 72.1 $\pm$ 0.5             & 74.3 $\pm$ 0.8          & 77.4                            \\
            MLDG               & 97.4 $\pm$ 0.2             & \textbf{65.2} $\pm$ 0.7    & 71.0 $\pm$ 1.4             & 75.3 $\pm$ 1.0          & 77.2                            \\
            CORAL              & 98.3 $\pm$ 0.1             & 66.1 $\pm$ 1.2             & 73.4 $\pm$ 0.3             & \textbf{77.5} $\pm$ 1.2 & \underline{78.8}                    \\
\bottomrule
        \end{tabular}
    }
    \label{table:Table_VLCS_training_full}%
\end{table*}

\begin{table*}[ht]%
    \caption{DomainBed benchmark, \textbf{VLCS full results for eave-one-domain-out cross-validation} model selection.}%
    \centering
    {
        \begin{tabular}{l|ccccc}
            \toprule
            \multirow{3}{*}{\textbf{Algorithm}} & \multicolumn{4}{c}{}                                                                                                                                                                                                                                                                                                                                                               \\           & \textbf{C}  & \textbf{L}   & \textbf{S}   & \textbf{V}      & \textbf{Avg}     \\
            \midrule
    ERM                  & 98.0 $\pm$ 0.4       & 62.6 $\pm$ 0.9       & 70.8 $\pm$ 1.9       & 77.5 $\pm$ 1.9       & 77.2                 \\
    IRM                  & \textbf{98.6} $\pm$ 0.3       & 66.0 $\pm$ 1.1       & 69.3 $\pm$ 0.9       & 71.5 $\pm$ 1.9       & 76.3                 \\
    GroupDRO             & 98.1 $\pm$ 0.3       &\underline{66.4} $\pm$ 0.9       & 71.0 $\pm$ 0.3       & 76.1 $\pm$ 1.4       & 77.9                 \\
    Mixup                & 98.4 $\pm$ 0.3       & 63.4 $\pm$ 0.7       & 72.9 $\pm$ 0.8       & 76.1 $\pm$ 1.2       & 77.7                 \\
    MLDG                 & \underline{98.5} $\pm$ 0.3       & 61.7 $\pm$ 1.2       & \textbf{73.6} $\pm$ 1.8       & 75.0 $\pm$ 0.8       & 77.2                 \\
    CORAL                & 96.9 $\pm$ 0.9       & 65.7 $\pm$ 1.2       & \underline{73.3} $\pm$ 0.7       & \textbf{78.7} $\pm$ 0.8       & \underline{78.7}                 \\
\bottomrule
        \end{tabular}
    }
    \label{table:Table_VLCS_validation_full}%
\end{table*}

\begin{table*}[ht]%
    \caption{\textbf{DomainBed benchmark}, VLCS full results for test-domain validation set (oracle) model selection method.}%
    \centering
    {
        \begin{tabular}{l|cccccc}
            \toprule
            \multirow{3}{*}{\textbf{Algorithm}} & \multicolumn{5}{c}{}                                                                                                                                                                                                                                                                                                                                                               \\           & \textbf{C}  & \textbf{L}   & \textbf{S}   & \textbf{V}      & \textbf{Avg}     \\
            \midrule
            ERM                & 97.6 $\pm$ 0.3             & 67.9 $\pm$ 0.7 & 70.9 $\pm$ 0.2             & 74.0 $\pm$ 0.6             & 77.6                           \\
            IRM                & 97.3 $\pm$ 0.2             & 66.7 $\pm$ 0.1             & 71.0 $\pm$ 2.3             & 72.8 $\pm$ 0.4             & 76.9                            \\
            GroupDRO           & 97.7 $\pm$ 0.2             & 65.9 $\pm$ 0.2             & 72.8 $\pm$ 0.8             & 73.4 $\pm$ 1.3             & 77.4                            \\
            Mixup              & 97.8 $\pm$ 0.4             & 67.2 $\pm$ 0.4             & 71.5 $\pm$ 0.2             & 75.7 $\pm$ 0.6             & 78.1                             \\
            MLDG               & 97.1 $\pm$ 0.5             & 66.6 $\pm$ 0.5             & 71.5 $\pm$ 0.1             & 75.0 $\pm$ 0.9             & 77.5                         \\
            CORAL              & 97.3 $\pm$ 0.2             & 67.5 $\pm$ 0.6             & 71.6 $\pm$ 0.6             & 74.5 $\pm$ 0.0             & 77.7                            \\
\bottomrule
        \end{tabular}
    }
    \label{table:Table_VLCS_oracle_full}%
\end{table*}

\clearpage 

\section{Prompts}
\label{subsec:prompts}

All prompts follow the same structure i.e., "a <class label>, <domain description>"
where the domain descriptions for PACS, OfficeHome, and DomainNet are as follows:

\subsection{PACS}

\begin{itemize}

\item  Photos: photorealistic, extremely detailed

\item Sketches: sketch drawing, black and white, less details

\item Cartoons: cartoon, cartoonish 

\item Art: art painting
\end{itemize}

\subsection{OfficeHome}

\begin{itemize}

\item  Clipart: Clipart, schematic, simplified 

\item Product: Product, Merchandise

\item Real: Real World, extremely detailed 

\item Art: art painting, art
\end{itemize}

\subsection{Domainnet}

\begin{itemize}

\item  Clipart: cartoon, cartoonish, drawing

 \item  Infograph: infographic, data visualization, poster 

\item   Real: photorealistic, extremely detailed  

\item   Painting: art painting  

\item  Quickdraw: extremely simple drawing, black and white 

\item  Sketch: sketch drawing, black and white, less details 

\item  Clipart: cartoon, cartoonish, drawing
\end{itemize}

\section{Code}
\label{subsec:Code}

To reproduce the DomainBed results, each class-specific dataset object inherits from either CDGA or CDGA$^*$ classes provided in this section. See the script provided in the section \ref{subsec:Code}.
\begin{lstlisting}[language=Python]
class CDGA(MultipleDomainDataset):
    def __init__(self, root, test_envs, augment, hparams):
        super().__init__()

        transform = transforms.Compose(
            [
                transforms.Resize((224, 224)),
                transforms.ToTensor(),
                transforms.Normalize(
                    mean=[0.485, 0.456, 0.406], std=[0.229, 0.224, 0.225]
                ),
            ]
        )

        augment_transform = transforms.Compose(
            [
                # transforms.Resize((224,224)),
                transforms.RandomResizedCrop(224, scale=(0.7, 1.0)),
                transforms.RandomHorizontalFlip(),
                transforms.ColorJitter(0.3, 0.3, 0.3, 0.3),
                transforms.RandomGrayscale(),
                transforms.ToTensor(),
                transforms.Normalize(
                    mean=[0.485, 0.456, 0.406], std=[0.229, 0.224, 0.225]
                ),
            ]
        )

        environments = [f.name for f in os.scandir(root) if f.is_dir()]
        environments = sorted(environments)

        self.datasets = []
        print(f"Test domains: {test_envs}")
        for i, environment in enumerate(environments):
            # Transformation
            if augment and (i not in test_envs):
                env_transform = augment_transform
            else:
                env_transform = transform

            path = os.path.join(root, environment)
            # Create list of generated subfolders for each distribution
            sub_environments = [f.name for f in os.scandir(path) if f.is_dir()]
            if i not in test_envs:
                # if we are in the training distribution combine folders that are not in the test distributions
                env_dataset = []
                for sub_env in sub_environments:
                    if all(environments[i] not in sub_env for i in test_envs):
                        print(f"Adding {sub_env} to {environment} for training")
                        env_dataset.append(
                            ImageFolder(
                                os.path.join(path, sub_env), transform=env_transform
                            )
                        )
                self.datasets.append(torch.utils.data.ConcatDataset(env_dataset))
            else:
                # if we are in the testing distribution just use the original data
                print(f"using {environment} for testing")
                self.datasets.append(
                    ImageFolder(
                        os.path.join(path, environment), transform=env_transform
                    )
                )
        self.input_shape = (
            3,
            224,
            224,
        )


class CDGA_star(MultipleDomainDataset):
    def __init__(self, root, test_envs, augment, hparams):
        super().__init__()

        transform = transforms.Compose(
            [
                transforms.Resize((224, 224)),
                transforms.ToTensor(),
                transforms.Normalize(
                    mean=[0.485, 0.456, 0.406], std=[0.229, 0.224, 0.225]
                ),
            ]
        )

        augment_transform = transforms.Compose(
            [
                # transforms.Resize((224,224)),
                transforms.RandomResizedCrop(224, scale=(0.7, 1.0)),
                transforms.RandomHorizontalFlip(),
                transforms.ColorJitter(0.3, 0.3, 0.3, 0.3),
                transforms.RandomGrayscale(),
                transforms.ToTensor(),
                transforms.Normalize(
                    mean=[0.485, 0.456, 0.406], std=[0.229, 0.224, 0.225]
                ),
            ]
        )

        environments = [f.name for f in os.scandir(root) if f.is_dir()]
        environments = sorted(environments)

        self.datasets = []
        print(f"Test domains: {test_envs}")
        for i, environment in enumerate(environments):
            if augment and (i not in test_envs):
                env_transform = augment_transform
            else:
                env_transform = transform
            path = os.path.join(root, environment)
            # create list of generated subfolders for each distribution
            sub_environments = [f.name for f in os.scandir(path) if f.is_dir()]
            if i not in test_envs:
                # if we are in the training distribution combine all the test folder except the original test data
                env_dataset = []
                for sub_env in sub_environments:
                    print(f"Adding {sub_env} to {environment} for training")
                    env_dataset.append(
                        ImageFolder(
                            os.path.join(path, sub_env), transform=env_transform
                        )
                    )
                self.datasets.append(torch.utils.data.ConcatDataset(env_dataset))
            else:
                # if we are in the testing distribution just use the original data
                print(f"using {environment} for testing")
                self.datasets.append(
                    ImageFolder(
                        os.path.join(path, environment), transform=env_transform
                    )
                )
        self.input_shape = (
            3,
            224,
            224,
        )


class G_PACS(CDGA):
    CHECKPOINT_FREQ = 300
    ENVIRONMENTS = ["A", "C", "P", "S"]
    num_classes = 7

    def __init__(self, root, test_envs, hparams):
        self.dir = os.path.join(root, "G_PACS/")
        super().__init__(self.dir, test_envs, hparams["data_augmentation"], hparams)
\end{lstlisting}

\end{document}